\pdfoutput=1
\documentclass[11pt]{article}

\usepackage[preprint]{acl}

\usepackage{times}
\usepackage{latexsym}

\usepackage[T1]{fontenc}
\usepackage[utf8]{inputenc}

\usepackage{microtype}

\usepackage{inconsolata}
\usepackage{graphicx} % DO NOT CHANGE THIS
\usepackage{caption} % DO NOT CHANGE THIS AND DO NOT ADD ANY OPTIONS TO IT

\usepackage{algorithm}
\usepackage{algorithmic}
\usepackage{amsmath}
\usepackage{amssymb}
\usepackage{subcaption}
\usepackage{float}
\usepackage{mathtools}
\usepackage{amsthm}
\usepackage{latexsym}
\usepackage{multirow}

\usepackage{soul}

\usepackage{colortbl}
\usepackage{booktabs}
\usepackage{scalerel}
\usepackage{tablefootnote}
\usepackage{footmisc}
\usepackage{ragged2e}
\usepackage{varioref}
\usepackage{cleveref}
\usepackage{arydshln}
\usepackage{tabularx}
\usepackage{xcolor}
\newcommand{\yesbut}{\textbf{\emph{YesBut}}}
\newcommand{\yesbutcolored}{\textbf{\emph{\textcolor{green}{Yes}\textcolor{red}{But}}}}

\title{\yesbutcolored: A High-Quality Annotated Multimodal Dataset for evaluating Satire Comprehension capability of Vision-Language Models}

\author{

Abhilash Nandy$^{\spadesuit}$ ~~~~
Yash Agarwal$^{\spadesuit}$ ~~~~
Ashish Patwa$^{\spadesuit}$ ~~~~
Millon Madhur Das$^{\spadesuit}$\\
\textbf{Aman Bansal}$^{\clubsuit}$ ~
\textbf{Ankit Raj}$^{\diamondsuit}$ ~
\textbf{Pawan Goyal}$^{\spadesuit}$ ~
\textbf{Niloy Ganguly}$^{\spadesuit}$\\
\texttt{nandyabhilash@kgpian.iitkgp.ac.in}
\\
$^\spadesuit$Indian Institute of Technology Kharagpur ~
$^{\clubsuit}$University of Massachusetts Amherst  \\
$^{\diamondsuit}$ Haldia Institute of Technology \\
}

\begin{document}
\maketitle
\begin{abstract}
Understanding satire and humor is a challenging task for even current Vision-Language models. In this paper, we propose the challenging tasks of \textit{Satirical Image Detection} (detecting whether an image is satirical), \textit{Understanding} (generating the reason behind the image being satirical), and \textit{Completion} (given one half of the image, selecting the other half from 2 given options, such that the complete image is satirical) and release a high-quality dataset \yesbut, consisting of 2547 images, 1084 satirical and 1463 non-satirical, containing different artistic styles, to evaluate those tasks. Each satirical image in the dataset depicts a normal scenario, along with a conflicting scenario which is funny or ironic. Despite the success of current Vision-Language Models on multimodal tasks such as Visual QA and Image Captioning, our benchmarking experiments show that such models perform poorly on the proposed tasks on the \yesbut\ Dataset in Zero-Shot Settings w.r.t both automated as well as human evaluation. Additionally, we release a dataset of 119 real, satirical photographs for further research\footnote{The dataset and code are available at \url{https://github.com/abhi1nandy2/yesbut_dataset}}.
% \footnote{Full \yesbut\ Dataset will be made public upon acceptance}.
\end{abstract}
\section{Introduction}

\begin{figure}[!t]
    \centering\includegraphics[width=0.35\textwidth]{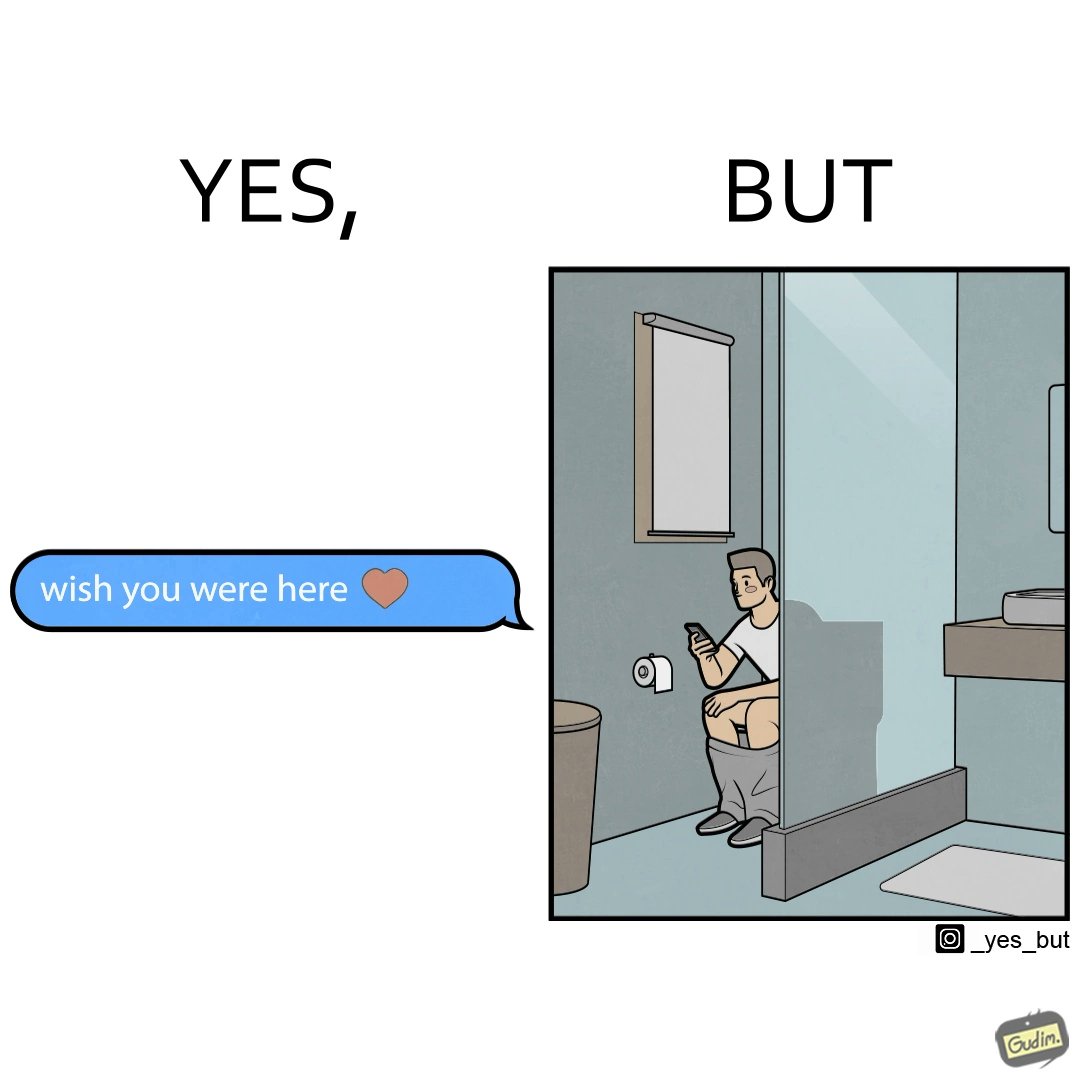}
    \caption{Satire conveyed through a social media image}
    \label{fig:yesbut_ex}
\end{figure}

Satire is a form of humor that uses irony or exaggeration to criticize or mock people, politics, or society. It serves as a powerful tool to highlight issues, provoke thought, and often encourages a critical perspective on the subject matter. Satirical images posted on social media often consist of conflicting scenarios to convey irony and humor.  Understanding such conflicting scenarios requires understanding interaction among entities and text (if any) within the image, along with commonsense knowledge and reasoning capabilities. Fig. \ref{fig:yesbut_ex} shows an example image conveying satire. The irony in the image is that the person is messaging someone a very heartfelt message on the mobile, while sitting on a toilet seat!

People convey humor on the internet and social media using images, GIFs, and videos. Previous studies have shown that memes \cite{meme1} and TV show Clips \cite{tvclips1} are prevalent means for expressing such humor. There have also been attempts at detecting \cite{hasan-etal-2019-ur, mustard, tanaka-etal-2022-learning} and describing \cite{hwang-shwartz-2023-memecap} multimodal satire and humor. However, very few works have \textit{simultaneously studied the detection, understanding, and comprehension of satirical situations in society in the multimodal setting}. %\pg{The above list has mixed references about generic works and computational works. Do you mean there is no computational work for this particular problem but is there for some related problems. If so, that needs to be made more explicit.} 

There has been a rise in the development of Vision-Language (VL) models \cite{llava, kosmos1, kosmos2, minigpt4, gpt4, gemini}. Such models have shown remarkable State-Of-The-Art (SOTA) performance on several downstream tasks such as Visual Question Answering and Image Captioning. Such models are pre-trained in a manner that images and text have shared embedding space, and that, images and their corresponding text descriptions have similar representations in that embedding space \cite{clip, siglip}.

In this paper, we investigate whether existing VL Models are able to decipher satire in images. To do so, we propose 3 benchmarking tasks - (1) {\sl Satirical Image Detection} - Given an image, classify the image as being satirical or not (2) {\sl Satirical Image Understanding} - Given a satirical image, describing in natural language why the image is satirical (3) {\sl Satirical Image Completion} - Given a part of the image, correctly select the remaining part of the image from 2 options. These tasks go beyond image recognition and language understanding, and are challenging, as understanding satire usually involves understanding the punchline corresponding to a sudden twist or a funny quip in a given situation \cite{ramachandran1998neurology}. For example, in Fig. \ref{fig:yesbut_ex}, the model needs to first comprehend the text "wish you were here", followed by understanding that the text was sent by a person sitting in the toilet from the image on the right, and then finally grasp the irony of the situation.

To evaluate the tasks, we collected a high-quality multimodal dataset \yesbut\ consisting of 1,084 satirical and 1,463 non-satirical images, where each image contains 2 sub-images with the same/different artistic styles. In each \textit{satirical image}, the left sub-image describes a scenario, and the right sub-image presents another scenario which either contradicts or pokes fun at the first scenario, creating an element of satire. Additionally, each such satirical sample is annotated to get the description of individual images inside the sample, as well as the overall description containing the punchline that conveys the satire.

We perform detailed evaluation on the satirical image detection, understanding, and completion tasks using recent VL models in zero-shot and zero-shot Chain-of-Thought (CoT) \cite{zero-shot-cot} settings (as we want to observe how well the models can decipher satire without the support of additional training/in-context examples). We observe that the task of satirical image detection is especially difficult. Also, even though Gemini performs the best in Satirical Image Understanding and Completion tasks, there is a significant scope for improvement in SOTA VL Models in understanding and comprehending satire in images in zero-shot scenarios. Also, for further research, we release an additional set of 119 diverse, real, satirical photographs. We infer that SOTA VL Models fail to perform well even on real photographs (\textit{\textbf{see Section \ref{appendix:intro} in Appendix}} for more details).

\section{Background}

\subsection{Satirical and Humor Datasets}

Previous works on satire and humor in NLP and Computer Vision mostly revolve around detecting satire in text \cite{rogoz-etal-2021-saroco} and multimodal scenarios \cite{li-etal-2020-multi-modal, ionescu2021fresada}, detecting humor in multimodal scenarios \cite{hasan-etal-2019-ur, mustard}, meme captioning \cite{hwang-shwartz-2023-memecap}, etc. However, no such work 
% \pg{Do we need a table discussing the difference of our dataset with other such benchmarks? That may actually be nice. Then, point 1 can become clear, so also point 2 can be put in context -- Expand your Table 1.} 
performs a comprehensive and simultaneous evaluation of satire and humor detection, understanding, and comprehension capabilities of VL Models in Multimodal Scenarios.

\subsection{Other Image Datasets}

The WHOOPS benchmark, introduced by \citet{whoops}, comprises unconventional images challenging commonsense expectations, both human-created and machine-generated, accompanied by corresponding textual descriptions. Specifically designed for tasks such as image captioning, image-text matching, visual question answering, and explanation generation, it provides a unique dataset for evaluating model performance in these domains. In contrast, our work performs a holistic evaluation of different SOTA VL Models on their ability to detect, understand, and comprehend satire in images.

\section{Our Annotation Pipeline}

\begin{figure*}[t]
    \centering
    \includegraphics[width=0.82\textwidth]{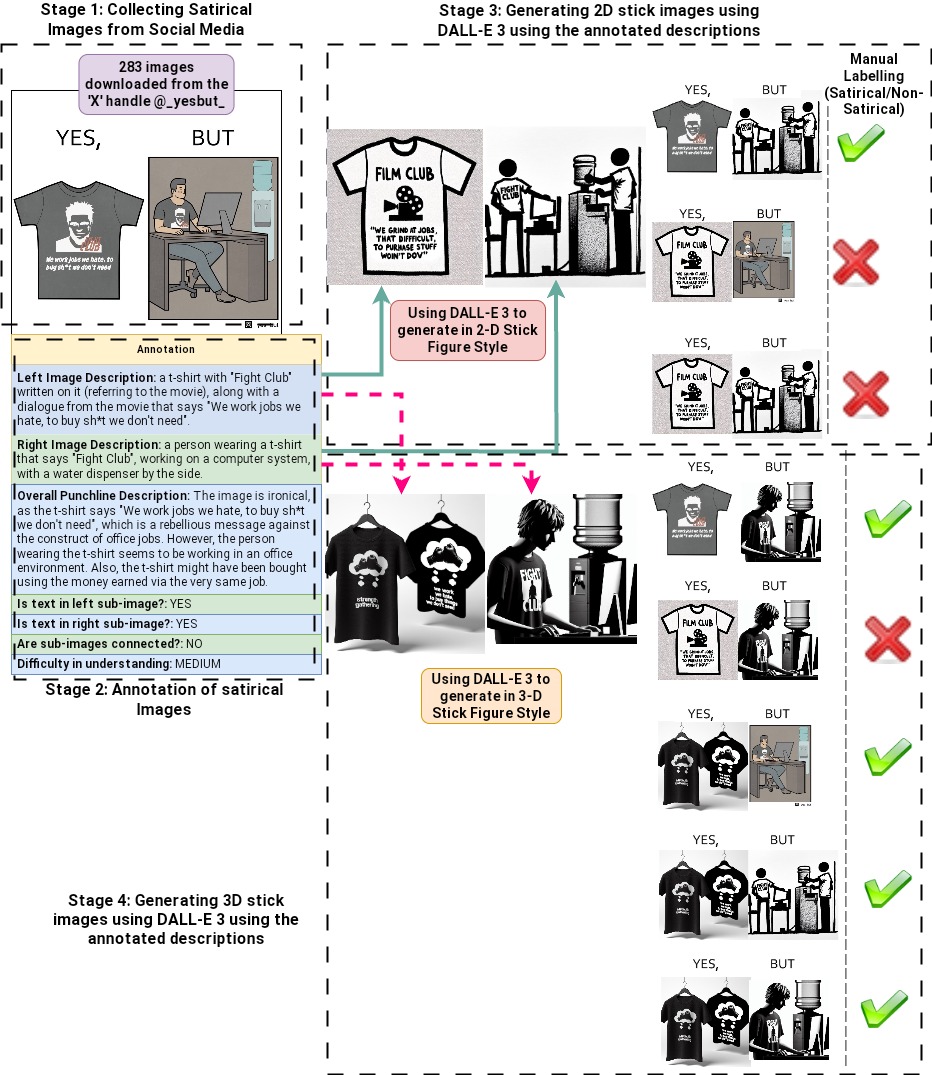}
    \caption{Our annotation Pipeline for \yesbut\ in 4 Stages - (1) Collecting Satirical Images from Social Media (2) Human Annotation of satirical images (3) Generating 2D stick images using DALL-E 3 and annotated descriptions (4) Generating 3D stick images using DALL-E 3 and annotated descriptions}
    \label{fig:annot_pipeline}
\end{figure*}

The entire data collection and annotation pipeline is shown in Fig. \ref{fig:annot_pipeline}. We curated a collection of annotated satirical and non-satirical images in this section in 4 stages.

\subsection{\textit{Stage 1}: Collecting Satirical Images from Social Media}
\label{sec:image_collection}

We manually downloaded images from the posts in `X' (erstwhile known as Twitter) handle \texttt{@\_yesbut\_} (with proper consent). We manually filtered 283 images that are satirical, and annotated them in the next stage. Each image contains two sub-images (which are colorized sketches), one on the left showing a normal scenario, while one on the right is ironical/pokes fun at the left sub-image.

\subsection{\textit{Stage 2}: Annotation of satirical images}
\label{sec:annotation}

Textual descriptions and certain categorical features of \textit{satirical} images were annotated using 5 annotators, all of whom met the qualification criteria of being undergraduate sophomore students or above, enrolled in English-medium colleges. Specifically, we collected the following features (these were given as annotator instructions) for every image - (1) Textual Description of the Left Sub-Image (2) Textual Description of the Right Sub-Image (3) Overall Textual Description which contains the punchline (4) A binary feature on whether the Left sub-Image contains any text (5) A binary feature on whether the Right sub-Image contains any text (6) A binary feature on whether the sub-images can be created by dividing a larger image using a vertical line as a separator (7) A categorical feature on how difficult the annotation was. This can have 3 possible values - `EASY' when the annotator does not need any additional help from the internet, `MEDIUM' when the annotator needs additional help from the internet to understand the overall description, and `HARD' when additional help from the internet is needed to write all the 3 aforementioned textual descriptions.

\begin{figure*}[t]
	\centering
	% \captionsetup{justification=centering}
	\subfloat[]{
		\centering 
\includegraphics[width=0.45\textwidth]{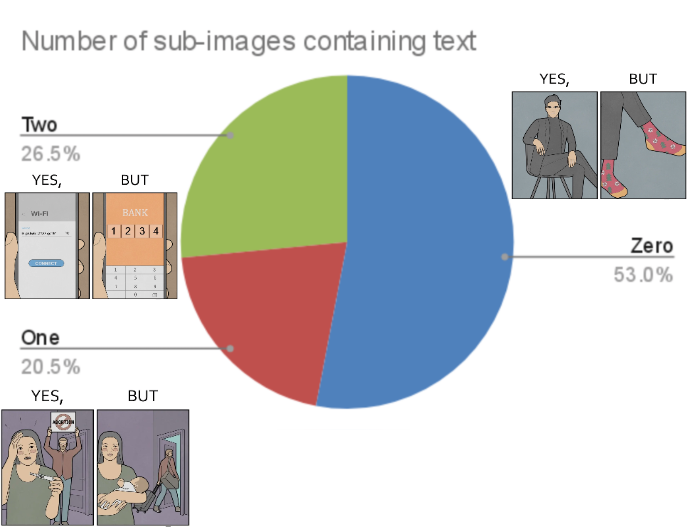} \label{text_count}}
% \vrule
	\subfloat[]{
		\centering 
\includegraphics[width=0.45\textwidth]{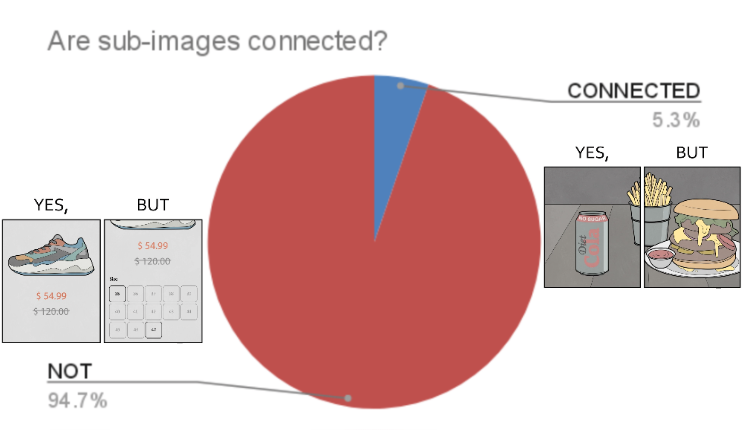} \label{conn}}

	\subfloat[]{
		\centering 
\includegraphics[width=0.4\textwidth]{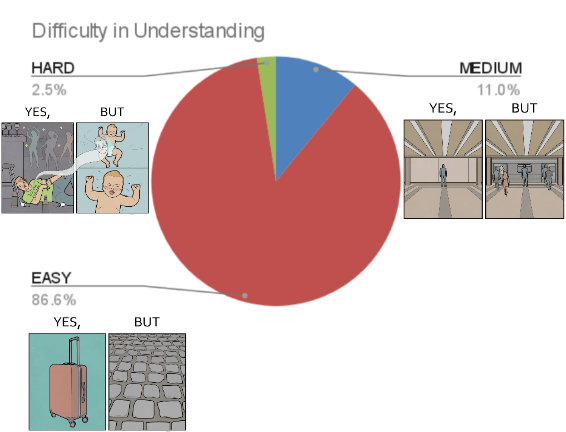} \label{diff_und}}
% \vrule
	\subfloat[]{
		\centering 
\includegraphics[width=0.5\textwidth]{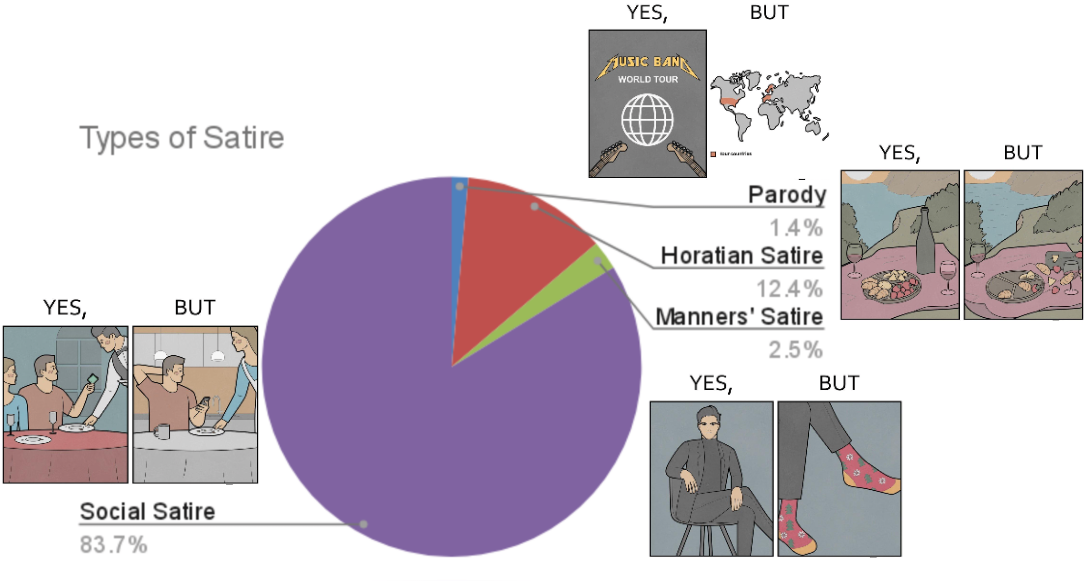} \label{satire}}

% \captionsetup{justification   = raggedright,
              % singlelinecheck = false}
	\caption{Distribution of the original 283 satirical images downloaded from Social Media based on different aspects of image content and annotated descriptions}
	%\vspace{-6mm}
	\label{fig:distr}
\end{figure*}

Fig. \ref{fig:distr} shows distribution of the 283 images based on different aspects of image content and annotated descriptions. We can see that - (1) from Fig. \ref{text_count}, more than half of the images have no text, which would make it difficult for the VL Models to understand those images due to absence of a text modality; (2) from Fig. \ref{conn}, more than 94\% of the images do not have connected sub-images, requiring the VL Models to understand the connection between the objects in the two sub-images; (3) from Fig. \ref{diff_und}, a significant 13.5\% of the (MEDIUM and HARD) images required annotators to refer to the internet to annotate the images, which makes the dataset challenging; (4) from Fig. \ref{satire}, the overall descriptions of the images containing the punchline were classified into 4 different types of satire using ChatGPT \cite{gpt35turbo}. Most of the images show Social Satire (it focuses on cultural trends, social conventions, and the absurdities of everyday life) and Horatian Satire (it aims to amuse rather than enrage, often using wit, irony, exaggeration to poke fun at societal norms and human folly).

\subsection{\textit{Stage 3}: Generating 2D stick images using DALL-E 3 on the annotated descriptions}

To increase the size and the diversity of the dataset, we use the DALL-E 3 \cite{dalle3} image generation model to generate synthetic sub-images using the annotated left and right sub-image descriptions (obtained in Stage 2). We use the following prompt - "Draw using stick figures (black silhouette against a white background) - <SUB-IMAGE DESCRIPTION>". Given the original sub-images, 3 new combinations of sub-images are obtained ([original left sub-image, generated right 2D stick sub-image], [generated left 2D stick sub-image, original right sub-image], [generated left 2D stick sub-image, generated right 2D stick sub-image]). We manually label each new combined image as satirical or non-satirical (details of this manual labelling is given \textit{\textbf{in Section \ref{annot_stage_3} of Appendix}}). %\pg{More details regarding this annotation are needed. Who were the annotators? Were they shown some examples? Annotation guidelines? Can be put in Appendix.}
At the end of the image generation followed by manual labelling, we end up adding 302 satirical and 547 non-satirical images. Each satirical image generated is assigned the same textual descriptions as the original image.

\subsection{\textit{Stage 4}: Generating 3D stick images using DALL-E 3 on the annotated descriptions}

Similar to Stage 3, we further increase the size and diversity using DALLE-3. We use the following prompt - "Draw using 3D black silhouettes against a white background - <SUB-IMAGE DESCRIPTION>". Given the original sub-images and the sub-images generated in Stage 3, 5 new combinations of sub-images are obtained ([original left sub-image, generated right 3D stick sub-image], [generated left 3D stick sub-image, original right sub-image], [generated left 2D stick sub-image, generated right 3D stick sub-image], [generated left 3D stick sub-image, generated right 2D stick sub-image], [generated left 3D stick sub-image, generated right 3D stick sub-image]). We manually label each new combined image as satirical or non-satirical. At the end of the image generation followed by manual labelling, we end up adding 499 satirical and 916 non-satirical images. Each satirical image generated is assigned the same textual descriptions as the original image.

\section{The \yesbut\ Dataset}

The \yesbut\ dataset has a total of 2,547 images, 1.084 of which are satirical, the rest 1,463 images being non-satirical. These images spread across 3 diverse artistic styles - colorized sketch, 2D stick figure, 3D stick figure.

% \usepackage{graphicx}
\begin{table}[H]
\resizebox{\columnwidth}{!}{%
\begin{tabular}{c|cccc}
\hline
\textbf{Dataset} & \textbf{\begin{tabular}[c]{@{}c@{}}Absence\\of text\\in image\end{tabular}} & \textbf{\begin{tabular}[c]{@{}c@{}}Presence of\\sub-images\end{tabular}} & \textbf{\begin{tabular}[c]{@{}c@{}}Presence of\\multiple\\ artistic styles\end{tabular}} & \textbf{\begin{tabular}[c]{@{}c@{}}Tasks relevant to\\multimodal humor\\comprehension\end{tabular}} \\ \hline
MemeCap         & 8.05\%                                                                       & 40.61\%                                                                   & 15.56\% &        \begin{tabular}[c]{@{}c@{}}Meme Captioning\end{tabular}                                                                         \\ \hline
MET-Meme         & 1.43\%                                                                       & 17.92\%                                                                   & 9.52\%       & \begin{tabular}[c]{@{}c@{}}Meme Understanding,\\Sentiment Analysis\end{tabular}                                                                          \\ \hline
\yesbut           & \textbf{53\%   }\tablefootnote{ignoring `YES', `BUT' Markers at the top of each image, as these markers are not related to the content of the image (see Fig. \ref{fig:yesbut_ex})}                                                                      & \textbf{100\%}                                                                     & \textbf{88.89\%}                                                           & \begin{tabular}[c]{@{}c@{}}Satirical Image\\Detection, \\Understanding,\\Completion\end{tabular}                     \\ \hline
\end{tabular}%
}
\caption{Statistics of the presence/absence of text, sub-images, and multiple image styles and tasks evaluated in prior datasets vs. \yesbut.}
\label{tab:dataset_stats}
\end{table}

Table \ref{tab:dataset_stats} compares \yesbut\ with multimodal satirical and humor datasets from prior art. We can see that \yesbut\ has a much larger fraction of images that (1) do not have text, (2) have sub-images, (3) have multiple artistic styles within the image, in comparison to MemeCap \cite{hwang-shwartz-2023-memecap} and MET-Meme \cite{metmeme} datasets. Lack of text and presence of multiple artistic styles across sub-images makes it challenging for the VL Models to comprehend satire in the images present in \yesbut. Additionally, the tasks in \yesbut\ ensure a more holistic evaluation of satire and humor compared to MemeCap and MET-Meme.

The satirical images cover several aspects of societal satire. To analyze this, we use topic modeling on the left and right sub-image descriptions using BERTopic \cite{bertopic}. We get 7 topics (each topic being an unordered set of representative words), which are further elaborated using ChatGPT to get intuitive descriptions for each topic (refer to \textbf{\textit{Section \ref{appendix:dataset} of Appendix}}).

\begin{figure}[ht]
    \centering
    \includegraphics[width=0.85\columnwidth]{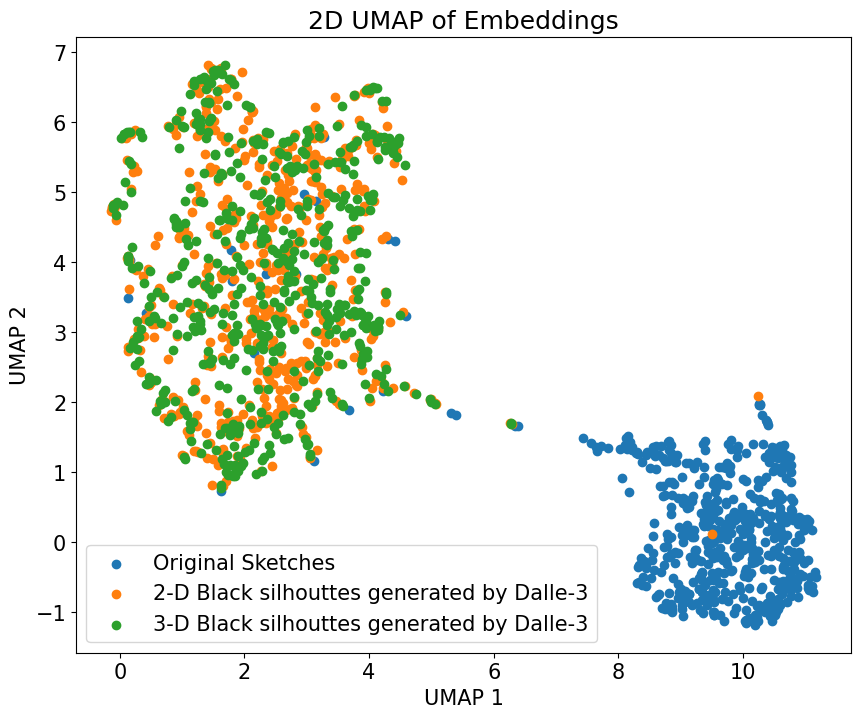}
    \caption{2D UMAP Representations of CLIP Image representations of \yesbut\ sub-images}
    \label{fig:image_embs}
\end{figure}

We further visualize the diversity of these sub-images by plotting the compressed 2D image representations obtained by applying UMAP \cite{umap} on the pre-trained CLIP \cite{clip} (MIT License) image representations in Fig. \ref{fig:image_embs}. The image samples are plotted in different colors based on their artistic style. The original 283 images are not very diverse. However, the generated images of the 2D and 3D stick figure styles are comparatively much more diverse and are semantically distant from the original images, even though they have the same sub-image descriptions. Hence, all the satirical images are highly diverse and cover various scenarios of societal satire.

\section{Experimental Setup}

We report the performance of various SOTA VL Models (described in Sec. \ref{sec:model_baselines}) for performance evaluation on the tasks (described in Sec. \ref{sec:tasks}) devised for the \yesbut\ Dataset. The evaluation setup and experimental results are described in Sec. \ref{sec:eval} and \ref{sec:results}, respectively.

\subsection{Models}
\label{sec:model_baselines}

\noindent \textbf{Gemini.} Gemini \cite{gemini} is a closed-source family of Large Multimodal Models (LMMs) from Google. The Gemini project comprises Ultra, Pro, and Nano variants, designed to excel in image and text comprehension. These models cater to diverse applications, from intricate reasoning tasks to memory-constrained on-device scenarios. Notably, the Gemini Ultra model demonstrates SOTA performance across 30/32 benchmarks. Furthermore, it outperforms existing models in all 20 multimodal benchmarks examined. The Gemini models showcase remarkable capabilities in cross-modal reasoning and language understanding. We leverage Gemini Pro Vision API for all tasks in our paper.

\noindent \textbf{GPT4.} GPT4 \cite{gpt4} is an advanced, closed-source multimodal model capable of processing both image, text inputs to generate coherent textual outputs. GPT4 demonstrates human-level proficiency across professional, academic benchmarks. It achieves commendable performance, ranking within the top 10\% of test takers in a simulated bar exam. Operating on an Autoregressive Transformer-based architecture \cite{vaswani2017attention}, GPT4 undergoes pre-training to predict subsequent tokens in a document. The subsequent post-training alignment enhances its performance in terms of factuality and adherence to desired behavior. We use \texttt{gpt-4-vision-preview} API for all tasks in our paper.

\noindent \textbf{LLaVA.} LLaVA  (Large Language and Vision Assistant), proposed by \citet{llava}, utilizes visual encoder from pre-trained CLIP \cite{clip} along with LLaMA \cite{llama} language model. The approach involves instruction tuning on visual instruction data assisted by GPT4 \cite{gpt4} for enhanced performance.

\noindent \textbf{MiniGPT4.} MiniGPT4 \cite{minigpt4} has frozen pre-trained language and vision components. It utilizes a singular projection layer to align visual and language features. Notably, it exhibits analogous capabilities to GPT4 in comprehending context. MiniGPT4 uses Vicuna \cite{vicuna} language model, built upon LLaMA-13B, demonstrating performance on par with ChatGPT. In the domain of vision, it integrates BLIP-2 \cite{blip2}, comprising CLIP ViT-G/14 \cite{clip} and a Q-Former \cite{qformer} architecture. Training MiniGPT4 encompasses diverse multimodal datasets, incorporating images from LAION \cite{laion}, Conceptual Captions \cite{sharma-etal-2018-conceptual}, and SBU \cite{sbu-captions}.

\noindent \textbf{Kosmos-2.} Equipped with a robust capability to comprehend diverse modalities, Kosmos-2 \cite{kosmos2} excels in undertaking an extensive array of tasks, ranging from zero-shot and few-shot scenarios to intricate multimodal chain-of-thought prompting situations. The model leverages textual instructions for enhanced comprehension of downstream tasks. In the context of chain-of-thought prompting, Kosmos-2 refines its approach by integrating grounding and referring capabilities, utilizing a structured format comprising text spans and bounding boxes as prompts. 
% Specifically, the format is represented as ``<p> text span </p><box><loc1><loc2></box>,'' where <loc1> and <loc2> denote location tokens, and <p>, </p>, <box>, and </box> serve as special boundary and text span tokens, respectively. 
This innovative approach enhances the model's effectiveness in generating coherent and contextually grounded responses, exemplifying the evolution from Kosmos-1 \cite{kosmos1}.

Table \ref{tab:vl-sizes} shows the number of parameters of the open-source VL Models\footnote{Compute Details are \textit{\textbf{in Section \ref{appendix:models} of Appendix}}}.

\begin{table}[H]
\centering
\resizebox{0.32\textwidth}{!}{%
\begin{tabular}{l|c}
\hline
 & \multicolumn{1}{c}{\textbf{\begin{tabular}[c]{@{}c@{}}Number\\of Parameters\\(In Billions)\end{tabular}}} \\ \hline
LLaVA \cite{llava} & 7 \\ 
Kosmos-2 \cite{kosmos2} & 1.66 \\
MiniGPT-4 \cite{minigpt4} & 15.7 \\
\hline
\end{tabular}
}
\caption{Number of Parameters - Open-Source Models}
\label{tab:vl-sizes}
\end{table}

\subsection{Tasks}
\label{sec:tasks}

We describe the tasks that are evaluated on the \yesbut\ Dataset -

\noindent \textit{Satirical Image Detection: }This is a binary classification task, where given an image, the model needs to predict whether the image is satirical or not. This task is carried out on all the 2547 images. Some example input images, along with input the text prompt used for all images is mentioned \textbf{\textit{in Section \ref{appendix:tasks} of Appendix.}}

\noindent \textit{Satirical Image Understanding: }Given a satirical image, we evaluate the model's satire understanding capability in images by (1) prompting the model to generate a textual description of each sub-image as input, using the prompt ``Describe the image''. (2) prompting the model to generate the punchline in the image using the following prompt (referred to as ``WHYFUNNY\_PROMPT'' hereafter) - ``Why is this image funny/satirical?''. 
% b. ``PUNCHLINE\_PROMPT'': ``This meme made me laugh. What is the punchline in this meme?''. 
This task is carried out on only the 1084 satirical images of the \yesbut\ Dataset.

\noindent \textit{Satirical Image Completion: }Given either the left or right sub-image having the style of a colorized sketch, the other sub-image needs to be chosen from two options, one having a 2D, and the other having a 3D stick figure style, such that the entire image so formed is meaningful and satirical. The options are curated based on existing satirical and non-satirical images from the \yesbut\ Dataset. We curate 150 such samples for evaluation. Some example input images, along with input the text prompt used for all images is mentioned \textbf{\textit{in Section \ref{appendix:tasks} of Appendix.}} %\pg{But aren't 2D and 3D images are paired? If so, one just needs to detect whether the current image is 2D/3D, and the task can be completed.}\abhi{constraint - colorized sketch sub-image in question, 2d, 3d in options}

\subsection{Evaluation Setup}
\label{sec:eval}

\noindent \textit{Satirical Image Detection: }We use Zero-Shot and Zero-Shot Chain-of-Thought (CoT) \cite{zero-shot-cot} setups for inference, and metrics used for binary classification such as Accuracy and F1-Score for evaluation. %Accuracy, Precision, Recall, and F1-Score for evaluation.

\noindent \textit{Satirical Image Understanding: }We use Zero-Shot setup for inference, and standard metrics for automatic evaluation of text generation-based tasks - lexical overlap metrics such as BLEU \cite{bleu}, ROUGE-L \cite{lin-2004-rouge}, and METEOR \cite{meteor}, and semantic similarity metrics such as BERTScore \cite{bertscore} to evaluate the image understanding capabilities of the images and corresponding sub-images (we also experiment with an image-based evaluation metric Polos \cite{polos}, whose results are shown \textit{\textbf{in Section \ref{appendix:results} of the Appendix}}). Additionally, we randomly sample 30 images (10 images from each obtained in Stage 2, Stage 3, Stage 4) along with their model-generated and human-written overall image descriptions. Each image description is human-evaluated based on the following (binary) criteria (adopted from \cite{hwang-shwartz-2023-memecap} and slightly changed\footnote{We do not use `Textual Completeness' from \citet{hwang-shwartz-2023-memecap}, as many images in \yesbut\ do not contain text} to better suit evaluation on \yesbut) - (1) \textbf{Correctness: }Is the image description correctly able to convey the satire the image wanted to convey? (2) \textbf{Appropriate Length: }Is the image description length appropriate for conveying the meaning (i.e. it is not too verbose)? (3) \textbf{Visual Completeness: }Does the image description describe all the important elements in the image?
(4) \textbf{Faithfulness: }Are all the elements of the image description supported by either the visual or text elements (i.e. there are no made-up elements)? - The annotation is carried out by 3 students in the lab\footnote{The annotators who annotated \yesbut\ were not a part of the human evaluation}, and the majority vote is taken for each image. %\pg{So, this is just binary scale?}\abhi{yes Sir, it's mentioned}

\noindent \textit{Satirical Image Completion: }We use Zero-Shot and Zero-Shot CoT setups for inference. and accuracy as the evaluation metric.

Note that we do not use In-Context Learning Setting for inference because this would make the tasks less challenging for the models. Also, we want to analyze how well VL models can comprehend satire on their own \textit{without any support from other exemplars}. 

\subsection{Results}
\label{sec:results}

\begin{figure*}[t]
    \centering
    \includegraphics[width=0.75\textwidth]{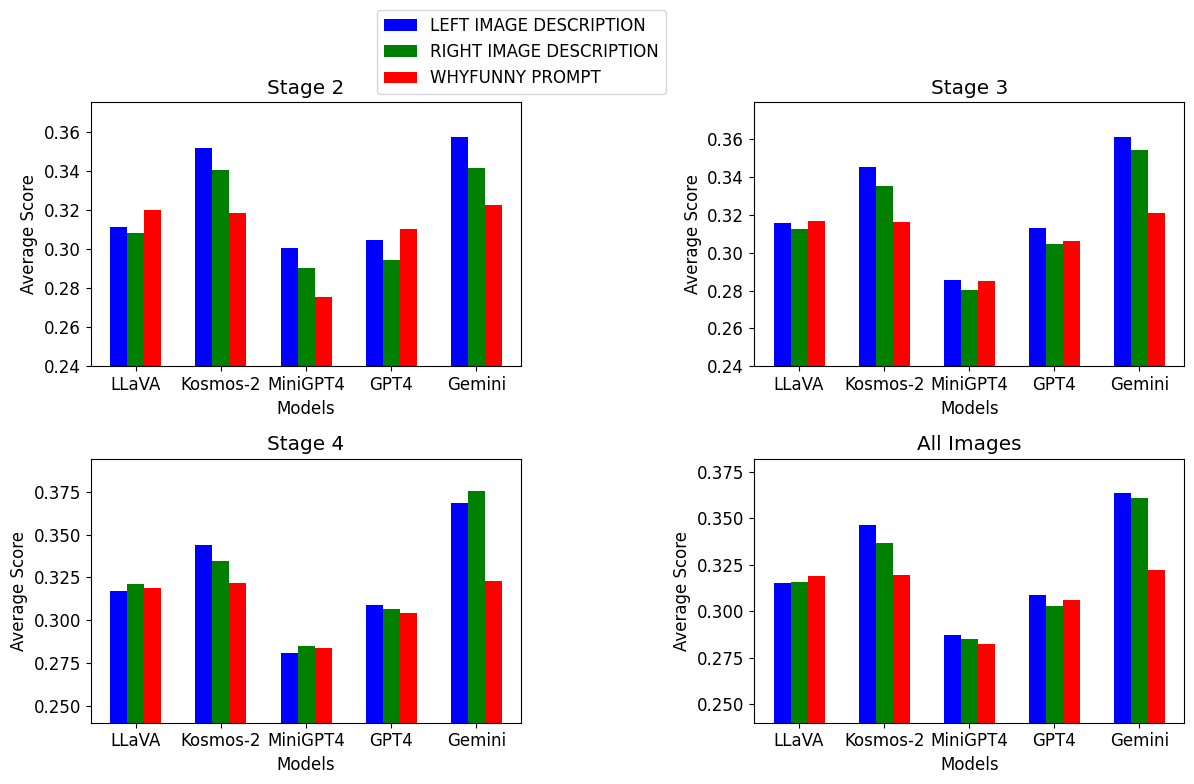}
    \caption{Evaluation of Satirical Image Understanding Capability using multiple VL models at different stages (Stages 2, 3, 4) of annotation of \yesbut, as well as, for all \yesbut\ images }
    \label{fig:satire-understanding}
\end{figure*}

\begin{table}[b]
\centering
\resizebox{0.32\textwidth}{!}{%
\begin{tabular}{l|cc}
\hline
 % & \multicolumn{1}{c}{\textbf{\begin{tabular}[c]{@{}c@{}}TEST\\ACC.\end{tabular}}} & \textbf{P} & \textbf{R} & \multicolumn{1}{c}{\textbf{\begin{tabular}[c]{@{}c@{}}F1\\SCORE\end{tabular}}} \\ \hline
 & \multicolumn{1}{c}{\textbf{\begin{tabular}[c]{@{}c@{}}TEST\\ACC.\end{tabular}}} & \multicolumn{1}{c}{\textbf{\begin{tabular}[c]{@{}c@{}}F1\\SCORE\end{tabular}}} \\ \hline
% LLaVA (0-shot) & 53.67 & 50.16 & 50.12 &  48.64\\ 
% LLaVA (0-shot, CoT) & 52.22 & 48.17 & 48.57 &   46.87 \\ 
LLaVA (0-shot) & 53.67 &  48.64\\ 
LLaVA (0-shot, CoT) & 52.22 &   46.87 \\ 
\hline

% Kosmos-2 (0-shot) & 42.56 & 42.56 & 100 &  59.71  \\
% Kosmos-2 (0-shot, CoT) & 56.97 & 47.94 & 12.92 &  20.35  \\
Kosmos-2 (0-shot) & 42.56 &  59.71  \\
Kosmos-2 (0-shot, CoT) & 56.97 &  20.35  \\
\hline

% MiniGPT4 (0-shot) & 48.29 & 59.13 & 42.31 &  49.33  \\
% MiniGPT4 (0-shot, CoT) & 48.88 & 61.53 & 42.98 &  50.61  \\
MiniGPT4 (0-shot) & 48.29 &  49.33  \\
MiniGPT4 (0-shot, CoT) & 48.88 &  50.61  \\
\hline

% GPT4 (0-shot) & 55.44 & 58.64 & 57.96 &  55.13  \\
% GPT4 (0-shot, CoT) & 48.29 & 61.27 & 54.29 &  42.32  \\
GPT4 (0-shot) & 55.44 &  55.13  \\
GPT4 (0-shot, CoT) & 48.29 &  42.32  \\
\hline

% Gemini (0-shot) & 50.82 & 57.72 & 55.16 &  48.29 \\
% Gemini (0-shot, CoT) & 46.36 & 58.46 & 52.51 & 38.93   \\
Gemini (0-shot) & 50.82 &  48.29 \\
Gemini (0-shot, CoT) & 46.36 & 38.93   \\
\hline

\end{tabular}
}
\caption{Evaluation of different VL models on the Satirical Image Detection task}
\label{tab:image-detection}
\end{table}

\noindent \textit{Satirical Image Detection: }Table \ref{tab:image-detection} shows the results of satirical image detection capability of VL Models on the \yesbut\ Dataset. 
% \pg{But aren't these all very poor for this binary task?}\abhi{even tho they are sota, see 5.1} 
We can infer that - (1) Kosmos-2 in zero-shot CoT and zero-shot settings give the best test accuracy and F1 Score respectively due to its superior visual grounding capabilities (2) Improvement in test accuracy and F1 Score due to CoT is seen only  in 2/5 and 1/5 models respectively, suggesting that SOTA VL Models are unable to properly reason/rationalize whether a given image has an element of satire in it (3) Both test accuracy and F1 Score do not cross 60\% for any SOTA VL Model, suggesting that there is a significant scope for improvement when it comes to detecting satire/humor in a given image.

\noindent \textit{Satirical Image Understanding: }Fig. \ref{fig:satire-understanding} shows the average value of the 4 automated metrics (discussed in Sec. \ref{sec:eval}) to evaluate satirical image understanding capability of VL Models at different stages of annotation of \yesbut\ (see Table \ref{tab:understanding} \textit{\textbf{in Section \ref{appendix:results} of Appendix}} to get individual values of the evaluation metrics, along with performance variation w.r.t annotation difficulty and presence of text in images). We observe that - (1) There is a reduction in the overall understanding capability (average metric corresponding to `WHYFUNNY PROMPT') of the majority of models in Stages 3 and 4 compared to Stage 2, as images in Stages 3 and 4 have different artistic styles in the same image, unlike Stage 2 (2) Kosmos-2 almost always performs better than other open-source models LLaVA and MiniGPT4, as Kosmos-2 has multimodal grounding and referring capabilities, which LLaVA and MiniGPT4 do not have (3) 4 out of 5 models do not understand the entire image better than sub-images within the image across the entire \yesbut\ Dataset. Gemini, Kosmos-2 encounter a huge drop in overall reasoning compared to sub-image reasoning, despite showing remarkable cross-modal reasoning and visual grounding capability respectively (4) MiniGPT4 gives the worst performance among all models due to restricted leverage of visual modality compared to textual modality, as stated in \citet{hwang-shwartz-2023-memecap}. (5) All average metric values (normalized between 0 and 1) are below 0.4, which shows that there is a lot of scope for improvement in satire understanding capability of SOTA VL Models. 

\begin{table}[H]
\centering
\resizebox{0.25\textwidth}{!}{%
\begin{tabular}{l|c}
\hline
 & \multicolumn{1}{c}{\textbf{\begin{tabular}[c]{@{}c@{}}TEST\\ ACC.\end{tabular}}} \\ \hline

LLaVA (0-shot) & 51.33 \\ 
LLaVA (0-shot, CoT) & 56.55 \\ 
\hline

Kosmos-2 (0-shot) & 54.67 \\
Kosmos-2 (0-shot, CoT) & 53.33 \\
\hline

MiniGPT4 (0-shot) & 40 \\
MiniGPT4 (0-shot, CoT) & 60.67 \\
\hline

GPT4 (0-shot) & 58.67 \\
GPT4 (0-shot, CoT) & 57.33 \\
\hline

Gemini (0-shot) & 61.11 \\
Gemini (0-shot, CoT) & 61.81 \\
\hline

\end{tabular}
}
\caption{Evaluation of different VL models on the Satirical Image Completion task}
\label{tab:image-completion}
\end{table}

\textit{\textbf{Figure \ref{fig:human-eval} in Section \ref{appendix:results} of Appendix}} compares the overall image descriptions generated by 5 SOTA Models with ones written by human annotators based on human evaluation (\textit{\textbf{see Table \ref{tab:appendix_yesbut_desc} in Section \ref{appendix:results} of Appendix}} for an example). We can see that Gemini and GPT4 perform satisfactorily among the 5 models. However, performance of the (aspect-wise) best model is 40, 43.33, 33.33, 36.66 points less compared to human-level performance on Correctness, Appropriate Length, Visual Completeness, and Faithfulness respectively.

\noindent \textit{Satirical Image Completion: }Table \ref{tab:image-completion} shows results of the satirical image completion task on \yesbut. We observe that - (1) CoT improves results in 3/5 models, as reasoning is needed to understand the relation between sub-images better (2) Among open-source models, improvement due to CoT is the highest for MiniGPT4, which is the largest open-source model in our study (see Table \ref{tab:vl-sizes}). 
% \pg{But in Table 3, miniGPT-4 had nearly no improvement here. Why?}\abhi{minigpt4 is the only model in table 3 which shows impr.} 
This is consistent with the observation of Zero-Shot CoT working better for larger models \cite{zero-shot-cot} (3) Gemini performs best in both zero-shot and zero-shot CoT settings among all the models.

\section{Summary and Conclusion}

We present \yesbut, a high-quality annotated multimodal dataset for Satire Comprehension Evaluation. Our work is one of the first to systematically benchmark multimodal Satire Comprehension ability of SOTA VL Models by proposing 3 non-trivial tasks of Satire Detection, Understanding, and Completion. We observe that SOTA VL Models struggle in those tasks, as \yesbut, unlike other benchmarks, contains images with sub-images having  different artistic styles and no text in most cases, making \yesbut\ a challenging multimodal dataset for satire detection and comprehension.
\section{Limitations}

\noindent \textbf{Subjectivity of annotations: }The annotation task involves utilizing background knowledge that may differ among annotators. Consequently, we manually reviewed the annotations to minimize the number of incorrect annotations in the dataset. However, some subjectivity still remains.

\noindent \textbf{Extension to languages other than English: }This work is in the English Language. However, we plan to extend our work to languages other than English.

\bibliography{custom}

\begin{thebibliography}{39}
\expandafter\ifx\csname natexlab\endcsname\relax\def\natexlab#1{#1}\fi

\bibitem[{Attardo et~al.(2003)Attardo, Eisterhold, Hay, and Poggi}]{tvclips1}
Salvatore Attardo, Jodi Eisterhold, Jennifer Hay, and Isabella Poggi. 2003.
\newblock \href {https://doi.org/doi:10.1515/humr.2003.012} {Multimodal markers of irony and sarcasm}.
\newblock \emph{HUMOR}, 16(2):243--260.

\bibitem[{Banerjee and Lavie(2005)}]{meteor}
Satanjeev Banerjee and Alon Lavie. 2005.
\newblock \href {https://aclanthology.org/W05-0909} {{METEOR}: An automatic metric for {MT} evaluation with improved correlation with human judgments}.
\newblock In \emph{Proceedings of the {ACL} Workshop on Intrinsic and Extrinsic Evaluation Measures for Machine Translation and/or Summarization}, pages 65--72, Ann Arbor, Michigan. Association for Computational Linguistics.

\bibitem[{Betker et~al.(2023)Betker, Goh, Jing, Brooks, Wang, Li, Ouyang, Zhuang, Lee, Guo et~al.}]{dalle3}
James Betker, Gabriel Goh, Li~Jing, Tim Brooks, Jianfeng Wang, Linjie Li, Long Ouyang, Juntang Zhuang, Joyce Lee, Yufei Guo, et~al. 2023.
\newblock Improving image generation with better captions.
\newblock \emph{Computer Science. https://cdn. openai. com/papers/dall-e-3. pdf}, 2(3):8.

\bibitem[{Bitton-Guetta et~al.(2023)Bitton-Guetta, Bitton, Hessel, Schmidt, Elovici, Stanovsky, and Schwartz}]{whoops}
Nitzan Bitton-Guetta, Yonatan Bitton, Jack Hessel, Ludwig Schmidt, Yuval Elovici, Gabriel Stanovsky, and Roy Schwartz. 2023.
\newblock \href {https://doi.org/10.48550/ARXIV.2303.07274} {Breaking common sense: Whoops! a vision-and-language benchmark of synthetic and compositional images}.

\bibitem[{Buchel(2012)}]{meme1}
Branislav Buchel. 2012.
\newblock Internet memes as means of communication.
\newblock \emph{Brno: Masaryk University}.

\bibitem[{Castro et~al.(2019)Castro, Hazarika, P{\'e}rez-Rosas, Zimmermann, Mihalcea, and Poria}]{mustard}
Santiago Castro, Devamanyu Hazarika, Ver{\'o}nica P{\'e}rez-Rosas, Roger Zimmermann, Rada Mihalcea, and Soujanya Poria. 2019.
\newblock Towards multimodal sarcasm detection (an \_obviously\_ perfect paper).
\newblock In \emph{Proceedings of the 57th Annual Meeting of the Association for Computational Linguistics (Volume 1: Long Papers)}, Florence, Italy. Association for Computational Linguistics.

\bibitem[{Chiang et~al.(2023)Chiang, Li, Lin, Sheng, Wu, Zhang, Zheng, Zhuang, Zhuang, Gonzalez, Stoica, and Xing}]{vicuna}
Wei-Lin Chiang, Zhuohan Li, Zi~Lin, Ying Sheng, Zhanghao Wu, Hao Zhang, Lianmin Zheng, Siyuan Zhuang, Yonghao Zhuang, Joseph~E. Gonzalez, Ion Stoica, and Eric~P. Xing. 2023.
\newblock \href {https://vicuna.lmsys.org} {Vicuna: An open-source chatbot impressing gpt-4 with 90\% chatgpt quality}.

\bibitem[{Grootendorst(2022)}]{bertopic}
Maarten Grootendorst. 2022.
\newblock Bertopic: Neural topic modeling with a class-based tf-idf procedure.
\newblock \emph{arXiv preprint arXiv:2203.05794}.

\bibitem[{Hasan et~al.(2019)Hasan, Rahman, Bagher~Zadeh, Zhong, Tanveer, Morency, and Hoque}]{hasan-etal-2019-ur}
Md~Kamrul Hasan, Wasifur Rahman, AmirAli Bagher~Zadeh, Jianyuan Zhong, Md~Iftekhar Tanveer, Louis-Philippe Morency, and Mohammed~(Ehsan) Hoque. 2019.
\newblock \href {https://doi.org/10.18653/v1/D19-1211} {{UR}-{FUNNY}: A multimodal language dataset for understanding humor}.
\newblock In \emph{Proceedings of the 2019 Conference on Empirical Methods in Natural Language Processing and the 9th International Joint Conference on Natural Language Processing (EMNLP-IJCNLP)}, pages 2046--2056, Hong Kong, China. Association for Computational Linguistics.

\bibitem[{Hayashi et~al.(2024)Hayashi, Sakai, Kamigaito, Hayashi, and Watanabe}]{hayashi2024artwork}
Kazuki Hayashi, Yusuke Sakai, Hidetaka Kamigaito, Katsuhiko Hayashi, and Taro Watanabe. 2024.
\newblock Artwork explanation in large-scale vision language models.
\newblock \emph{arXiv preprint arXiv:2403.00068}.

\bibitem[{Huang et~al.(2023)Huang, Dong, Wang, Hao, Singhal, Ma, Lv, Cui, Mohammed, Liu et~al.}]{kosmos1}
Shaohan Huang, Li~Dong, Wenhui Wang, Yaru Hao, Saksham Singhal, Shuming Ma, Tengchao Lv, Lei Cui, Owais~Khan Mohammed, Qiang Liu, et~al. 2023.
\newblock Language is not all you need: Aligning perception with language models.
\newblock \emph{arXiv preprint arXiv:2302.14045}.

\bibitem[{Hwang and Shwartz(2023)}]{hwang-shwartz-2023-memecap}
EunJeong Hwang and Vered Shwartz. 2023.
\newblock \href {https://doi.org/10.18653/v1/2023.emnlp-main.89} {{M}eme{C}ap: A dataset for captioning and interpreting memes}.
\newblock In \emph{Proceedings of the 2023 Conference on Empirical Methods in Natural Language Processing}, pages 1433--1445, Singapore. Association for Computational Linguistics.

\bibitem[{Ionescu and Chifu(2021)}]{ionescu2021fresada}
Radu~Tudor Ionescu and Adrian~Gabriel Chifu. 2021.
\newblock Fresada: A french satire data set for cross-domain satire detection.
\newblock In \emph{2021 International Joint Conference on Neural Networks (IJCNN)}, pages 1--8. IEEE.

\bibitem[{Kojima et~al.(2022)Kojima, Gu, Reid, Matsuo, and Iwasawa}]{zero-shot-cot}
Takeshi Kojima, Shixiang~(Shane) Gu, Machel Reid, Yutaka Matsuo, and Yusuke Iwasawa. 2022.
\newblock \href {https://proceedings.neurips.cc/paper_files/paper/2022/file/8bb0d291acd4acf06ef112099c16f326-Paper-Conference.pdf} {Large language models are zero-shot reasoners}.
\newblock In \emph{Advances in Neural Information Processing Systems}, volume~35, pages 22199--22213. Curran Associates, Inc.

\bibitem[{Li et~al.(2023)Li, Li, Savarese, and Hoi}]{blip2}
Junnan Li, Dongxu Li, Silvio Savarese, and Steven Hoi. 2023.
\newblock \href {http://arxiv.org/abs/2301.12597} {Blip-2: Bootstrapping language-image pre-training with frozen image encoders and large language models}.

\bibitem[{Li et~al.(2020)Li, Levi, Hosseini, and Broniatowski}]{li-etal-2020-multi-modal}
Lily Li, Or~Levi, Pedram Hosseini, and David Broniatowski. 2020.
\newblock \href {https://aclanthology.org/2020.nlp4if-1.4} {A multi-modal method for satire detection using textual and visual cues}.
\newblock In \emph{Proceedings of the 3rd NLP4IF Workshop on NLP for Internet Freedom: Censorship, Disinformation, and Propaganda}, pages 33--38, Barcelona, Spain (Online). International Committee on Computational Linguistics (ICCL).

\bibitem[{Lin(2004)}]{lin-2004-rouge}
Chin-Yew Lin. 2004.
\newblock \href {https://aclanthology.org/W04-1013} {{ROUGE}: A package for automatic evaluation of summaries}.
\newblock In \emph{Text Summarization Branches Out}, pages 74--81, Barcelona, Spain. Association for Computational Linguistics.

\bibitem[{Liu et~al.(2023)Liu, Li, Wu, and Lee}]{llava}
Haotian Liu, Chunyuan Li, Qingyang Wu, and Yong~Jae Lee. 2023.
\newblock \href {http://arxiv.org/abs/2304.08485} {Visual instruction tuning}.

\bibitem[{McInnes et~al.(2018)McInnes, Healy, Saul, and Gro{\ss}berger}]{umap}
Leland McInnes, John Healy, Nathaniel Saul, and Lukas Gro{\ss}berger. 2018.
\newblock Umap: Uniform manifold approximation and projection.
\newblock \emph{Journal of Open Source Software}, 3(29).

\bibitem[{OpenAI(2021)}]{gpt35turbo}
OpenAI. 2021.
\newblock \href {https://platform.openai.com/docs/models/gpt-3-5} {Gpt-3.5 turbo documentation}.

\bibitem[{OpenAI(2023)}]{gpt4}
OpenAI. 2023.
\newblock \href {http://arxiv.org/abs/2303.08774} {Gpt-4 technical report}.

\bibitem[{Ordonez et~al.(2011)Ordonez, Kulkarni, and Berg}]{sbu-captions}
Vicente Ordonez, Girish Kulkarni, and Tamara Berg. 2011.
\newblock \href {https://proceedings.neurips.cc/paper/2011/file/5dd9db5e033da9c6fb5ba83c7a7ebea9-Paper.pdf} {Im2text: Describing images using 1 million captioned photographs}.
\newblock In \emph{Advances in Neural Information Processing Systems}, volume~24. Curran Associates, Inc.

\bibitem[{Papineni et~al.(2002)Papineni, Roukos, Ward, and Zhu}]{bleu}
Kishore Papineni, Salim Roukos, Todd Ward, and Wei-Jing Zhu. 2002.
\newblock \href {https://doi.org/10.3115/1073083.1073135} {{B}leu: a method for automatic evaluation of machine translation}.
\newblock In \emph{Proceedings of the 40th Annual Meeting of the Association for Computational Linguistics}, pages 311--318, Philadelphia, Pennsylvania, USA. Association for Computational Linguistics.

\bibitem[{Peng et~al.(2023)Peng, Wang, Dong, Hao, Huang, Ma, and Wei}]{kosmos2}
Zhiliang Peng, Wenhui Wang, Li~Dong, Yaru Hao, Shaohan Huang, Shuming Ma, and Furu Wei. 2023.
\newblock Kosmos-2: Grounding multimodal large language models to the world.
\newblock \emph{arXiv preprint arXiv:2306.14824}.

\bibitem[{Radford et~al.(2021)Radford, Kim, Hallacy, Ramesh, Goh, Agarwal, Sastry, Askell, Mishkin, Clark et~al.}]{clip}
Alec Radford, Jong~Wook Kim, Chris Hallacy, Aditya Ramesh, Gabriel Goh, Sandhini Agarwal, Girish Sastry, Amanda Askell, Pamela Mishkin, Jack Clark, et~al. 2021.
\newblock Learning transferable visual models from natural language supervision.
\newblock In \emph{International conference on machine learning}, pages 8748--8763. PMLR.

\bibitem[{Ramachandran(1998)}]{ramachandran1998neurology}
Vilayanur~S Ramachandran. 1998.
\newblock The neurology and evolution of humor, laughter, and smiling: the false alarm theory.
\newblock \emph{Medical hypotheses}, 51(4):351--354.

\bibitem[{Rogoz et~al.(2021)Rogoz, Mihaela, and Ionescu}]{rogoz-etal-2021-saroco}
Ana-Cristina Rogoz, Gaman Mihaela, and Radu~Tudor Ionescu. 2021.
\newblock \href {https://doi.org/10.18653/v1/2021.acl-short.136} {{S}a{R}o{C}o: Detecting satire in a novel {R}omanian corpus of news articles}.
\newblock In \emph{Proceedings of the 59th Annual Meeting of the Association for Computational Linguistics and the 11th International Joint Conference on Natural Language Processing (Volume 2: Short Papers)}, pages 1073--1079, Online. Association for Computational Linguistics.

\bibitem[{Schuhmann et~al.(2022)Schuhmann, Beaumont, Vencu, Gordon, Wightman, Cherti, Coombes, Katta, Mullis, Wortsman, Schramowski, Kundurthy, Crowson, Schmidt, Kaczmarczyk, and Jitsev}]{laion}
Christoph Schuhmann, Romain Beaumont, Richard Vencu, Cade Gordon, Ross Wightman, Mehdi Cherti, Theo Coombes, Aarush Katta, Clayton Mullis, Mitchell Wortsman, Patrick Schramowski, Srivatsa Kundurthy, Katherine Crowson, Ludwig Schmidt, Robert Kaczmarczyk, and Jenia Jitsev. 2022.
\newblock \href {http://arxiv.org/abs/2210.08402} {Laion-5b: An open large-scale dataset for training next generation image-text models}.

\bibitem[{Sharma et~al.(2018)Sharma, Ding, Goodman, and Soricut}]{sharma-etal-2018-conceptual}
Piyush Sharma, Nan Ding, Sebastian Goodman, and Radu Soricut. 2018.
\newblock \href {https://doi.org/10.18653/v1/P18-1238} {Conceptual captions: A cleaned, hypernymed, image alt-text dataset for automatic image captioning}.
\newblock In \emph{Proceedings of the 56th Annual Meeting of the Association for Computational Linguistics (Volume 1: Long Papers)}, pages 2556--2565, Melbourne, Australia. Association for Computational Linguistics.

\bibitem[{Tanaka et~al.(2022)Tanaka, Yamane, Mori, Mukuta, and Harada}]{tanaka-etal-2022-learning}
Kohtaro Tanaka, Hiroaki Yamane, Yusuke Mori, Yusuke Mukuta, and Tatsuya Harada. 2022.
\newblock \href {https://aclanthology.org/2022.cai-1.9} {Learning to evaluate humor in memes based on the incongruity theory}.
\newblock In \emph{Proceedings of the Second Workshop on When Creative AI Meets Conversational AI}, pages 81--93, Gyeongju, Republic of Korea. Association for Computational Linguistics.

\bibitem[{Team(2023)}]{gemini}
Gemini Team. 2023.
\newblock \href {http://arxiv.org/abs/2312.11805} {Gemini: A family of highly capable multimodal models}.

\bibitem[{Touvron et~al.(2023)Touvron, Lavril, Izacard, Martinet, Lachaux, Lacroix, Rozière, Goyal, Hambro, Azhar, Rodriguez, Joulin, Grave, and Lample}]{llama}
Hugo Touvron, Thibaut Lavril, Gautier Izacard, Xavier Martinet, Marie-Anne Lachaux, Timothée Lacroix, Baptiste Rozière, Naman Goyal, Eric Hambro, Faisal Azhar, Aurelien Rodriguez, Armand Joulin, Edouard Grave, and Guillaume Lample. 2023.
\newblock \href {http://arxiv.org/abs/2302.13971} {Llama: Open and efficient foundation language models}.

\bibitem[{Vaswani et~al.(2017)Vaswani, Shazeer, Parmar, Uszkoreit, Jones, Gomez, Kaiser, and Polosukhin}]{vaswani2017attention}
Ashish Vaswani, Noam Shazeer, Niki Parmar, Jakob Uszkoreit, Llion Jones, Aidan~N Gomez, {\L}ukasz Kaiser, and Illia Polosukhin. 2017.
\newblock Attention is all you need.
\newblock \emph{Advances in neural information processing systems}, 30.

\bibitem[{Wada et~al.(2024)Wada, Kaneda, Saito, and Sugiura}]{polos}
Yuiga Wada, Kanta Kaneda, Daichi Saito, and Komei Sugiura. 2024.
\newblock {Polos: Multimodal Metric Learning from Human Feedback for Image Captioning}.
\newblock In \emph{Proceedings of the IEEE/CVF Conference on Computer Vision and Pattern Recognition}.

\bibitem[{Xu et~al.(2022)Xu, Li, Zheng, Naseriparsa, Zhao, Lin, and Xia}]{metmeme}
Bo~Xu, Tingting Li, Junzhe Zheng, Mehdi Naseriparsa, Zhehuan Zhao, Hongfei Lin, and Feng Xia. 2022.
\newblock \href {https://doi.org/10.1145/3477495.3532019} {Met-meme: A multimodal meme dataset rich in metaphors}.
\newblock In \emph{Proceedings of the 45th International ACM SIGIR Conference on Research and Development in Information Retrieval}, SIGIR '22, page 2887–2899, New York, NY, USA. Association for Computing Machinery.

\bibitem[{Zhai et~al.(2023)Zhai, Mustafa, Kolesnikov, and Beyer}]{siglip}
Xiaohua Zhai, Basil Mustafa, Alexander Kolesnikov, and Lucas Beyer. 2023.
\newblock Sigmoid loss for language image pre-training.
\newblock \emph{arXiv preprint arXiv:2303.15343}.

\bibitem[{Zhang et~al.(2024)Zhang, Zhang, Xu, and Tao}]{qformer}
Qiming Zhang, Jing Zhang, Yufei Xu, and Dacheng Tao. 2024.
\newblock Vision transformer with quadrangle attention.
\newblock \emph{IEEE Transactions on Pattern Analysis and Machine Intelligence}.

\bibitem[{Zhang* et~al.(2020)Zhang*, Kishore*, Wu*, Weinberger, and Artzi}]{bertscore}
Tianyi Zhang*, Varsha Kishore*, Felix Wu*, Kilian~Q. Weinberger, and Yoav Artzi. 2020.
\newblock \href {https://openreview.net/forum?id=SkeHuCVFDr} {Bertscore: Evaluating text generation with bert}.
\newblock In \emph{International Conference on Learning Representations}.

\bibitem[{Zhu et~al.(2023)Zhu, Chen, Shen, Li, and Elhoseiny}]{minigpt4}
Deyao Zhu, Jun Chen, Xiaoqian Shen, Xiang Li, and Mohamed Elhoseiny. 2023.
\newblock \href {http://arxiv.org/abs/2304.10592} {Minigpt-4: Enhancing vision-language understanding with advanced large language models}.

\end{thebibliography}

\appendix

\section*{Appendix}
\label{sec:appendix}

The Appendix mirrors the sectional structure of the main paper, placing supplementary material for each section in its corresponding appendix section for easy reference. If some sections or subsections lack additional material, only their titles are listed.

\section{Introduction}
\label{appendix:intro}

\begin{figure}[ht]
    \centering
    \includegraphics[width=\columnwidth]{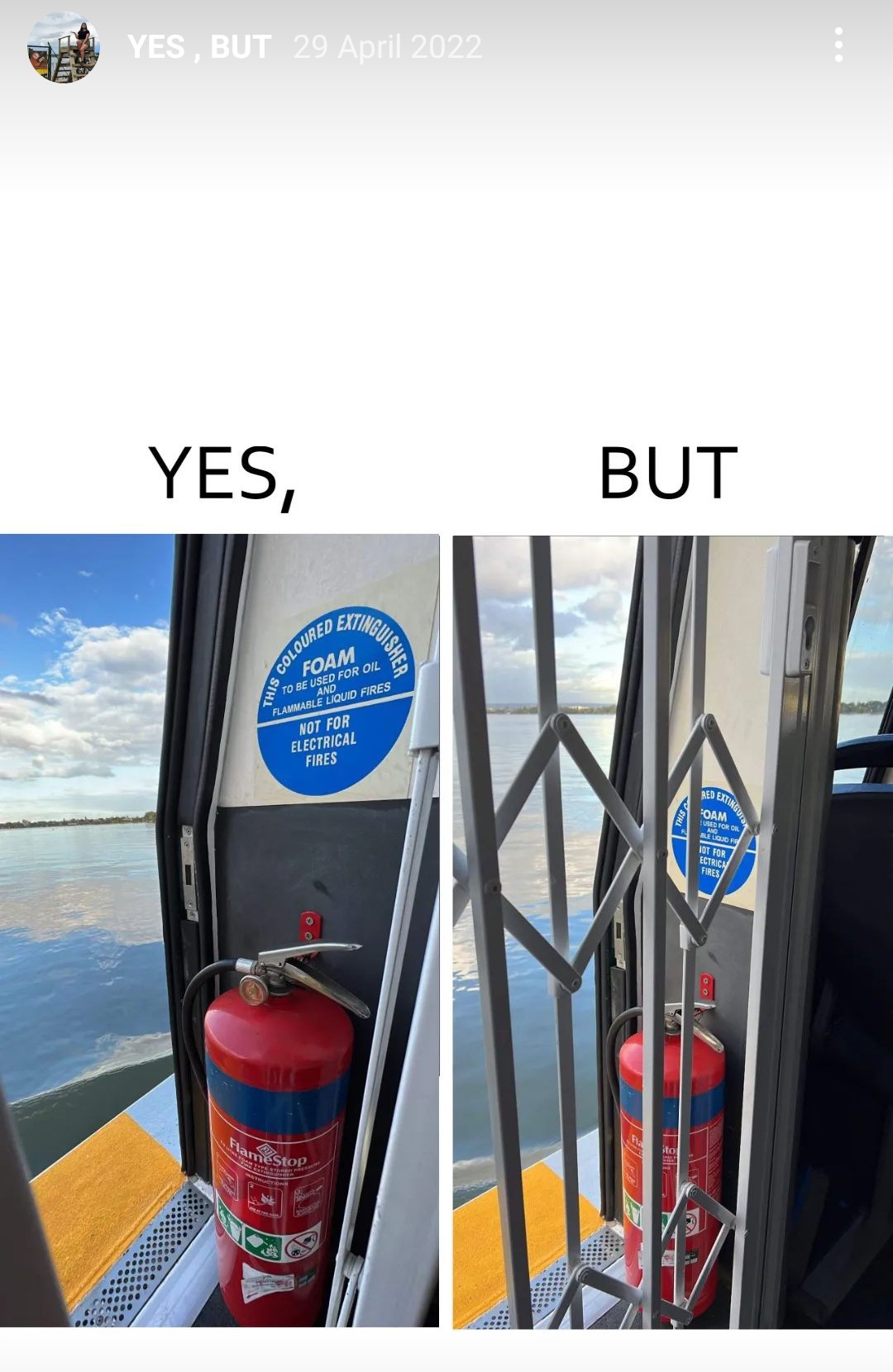}
    \caption{Example of a real photograph following the `Yes, But' Theme}
    \label{fig:real_photo}
\end{figure}

\noindent \textbf{Dataset of real, satirical images: }We collected a dataset of 119 images containing irony, satire from instagram posts by different users, who resort to using “Yes, But” theme over real photos (e.g. see Figure \ref{fig:real_photo}). We perform following 2 tasks on these images - (1) Satirical Image Detection, where we report detection accuracy, as all images have ground truth of “Satirical” (2) Satirical Image Understanding, where we use the WHYFUNNY text prompt and the image as input to the VL Models. The output is evaluated using human evaluation, where the annotator needs to answer whether model-generated text correctly describes satire in the image, and the corresponding accuracy for each VL Model is reported. The results are shown in Table \ref{tab:real-photos}. We can infer that 3 out of 5 models give less than satisfactory performance on Detection, and all models give an accuracy of less than 50\% on Image Understanding. Hence, even on real photographs, SOTA VL Models fail to perform well.

% \usepackage{graphicx}
\begin{table}[H]
\centering
\resizebox{0.8\columnwidth}{!}{%
\begin{tabular}{l|ll}
\hline
\textbf{Model} & \textbf{Detection} & \textbf{Understanding} \\ \hline
LLaVA          & 18.48              & 22.69                  \\
Kosmos-2       & 66.39              & 10.92                  \\
MiniGPT4       & 52.94              & 10.92                  \\
GPT4           & 93.27              & 46.22                  \\
Gemini         & 80.67              & 19.33                  \\ \hline
\end{tabular}%
}
\caption{Performance of different SOTA VL Models on Satirical Detection and Understanding Tasks on real photographs}
\label{tab:real-photos}
\end{table}

\section{Background}
\subsection{Satirical and Humor Datasets}
\subsection{Other Image Datasets}

\section{Our Annotation Pipeline}

\subsection{\textit{Stage 1}: Collecting Satirical Images from Social Media}
\subsection{\textit{Stage 2}: Annotation of Satirical Images}
\subsection{\textit{Stage 3}: Generating 2D stick images using DALL-E 3 on the annotated descriptions}
\label{annot_stage_3}

\textbf{Details of the manual labelling: }The manual labelling of whether an image with one or more generated sub-images is satirical or not is carried out by a graduate student in our lab. The annotator was given 10 satirical and 10 non-satirical images prior to the manual labelling to provide assistance for the labelling.

\subsection{\textit{Stage 4}: Generating 3D stick images using DALL-E 3 on the annotated descriptions}

\section{The \yesbut\ Dataset}
\label{appendix:dataset}

\noindent \textbf{Topics obtained after topic-modelling on the left and right sub-image descriptions of satirical images in \yesbut, along with topic descriptions from ChatGPT -}

\begin{itemize}
    \item \textit{gate\_shorts\_step\_allowed\_person}: Likely related to airport security procedures or access control systems, involving individuals wearing shorts being allowed to proceed through a gate or checkpoint.
    \item \textit{phone\_screen\_mobile\_smartphone\_person}: Refers to activities or interactions involving individuals using their smartphones, possibly related to mobile technology, communication, or digital engagement.

\item \textit{woman\_image\_shows\_saying\_text}: Implies content featuring women in images, possibly conveying messages or text, suggesting contexts such as advertisements, social media posts, or presentations.

\item \textit{plate\_table\_food\_box\_cup}: Indicates elements commonly found in dining or food service settings, encompassing plates, tables, various food items, boxes, and cups, suggesting scenarios like restaurants or meal preparation.

\item \textit{person\_wearing\_hair\_tattoos\_pants}: Describes characteristics of individuals including their clothing choices (pants), hairstyles, and tattoos, likely relevant in contexts such as fashion, identity expression, or cultural representations.

\item \textit{car\_light\_traffic\_road\_image}: Depicts scenes involving cars, traffic conditions, and roads, possibly associated with transportation, urban environments, or traffic management, often visualized through images.

\item \textit{dog\_hole\_cat\_two\_throw}: Suggests actions or scenarios involving dogs, cats, and interactions such as throwing, possibly indicating playful or behavioral aspects of these animals, possibly related to pet ownership or animal behavior studies.
\end{itemize}

\section{Experimental Setup}

\subsection{Models}
\label{appendix:models}

\noindent \textbf{Compute Details: }We use an NVIDIA A40 GPU for experiments using the open-source models. The inference time per sample on the GPU for the Satirical Image Detection, Understanding and Completion Tasks for the open-source models go upto around 10 seconds, 1 minute, and 10 seconds respectively.

\subsection{Tasks}
\label{appendix:tasks}
\textbf{Text Prompt for \textit{Satirical Image Detection}: }

\texttt{You are an AI expert in detecting humour or satire. User gives you an image, and you have to make a choice "Y" or "N".
Instructions: Users image has 2 halves called yes and but, and the combination of those might make no sense at all, or be extremely funny. Your job is to find out which one it is and output Y if its EXTREMELY funny and N for otherwise.
Output format: one character, exactly either "Y" or "N"}

\noindent \textbf{Example Image Inputs for \textit{Satirical Image Detection}: }

\begin{figure}[H]
    \centering
    \includegraphics[width=\columnwidth]{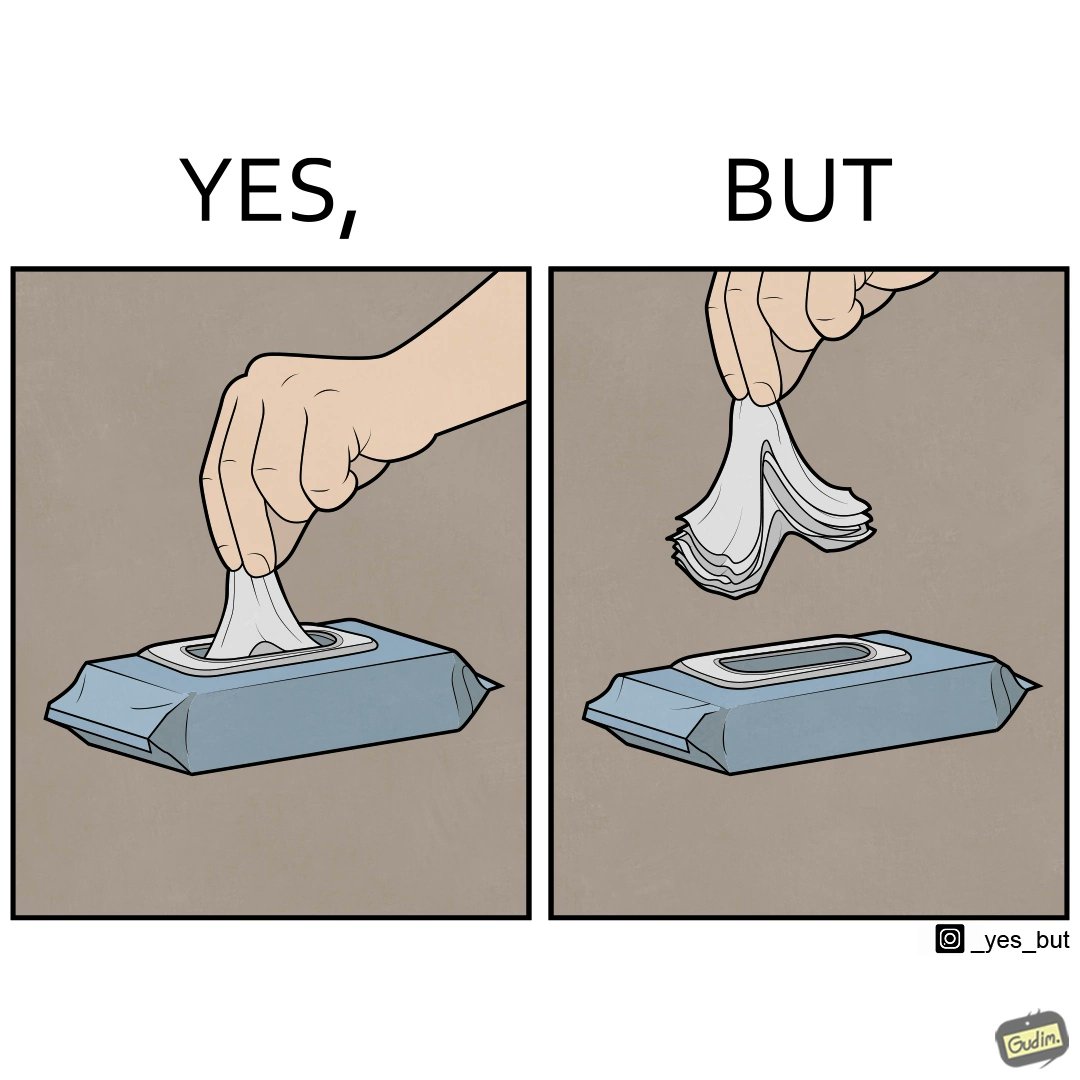}
    \caption{Example of a Satirical Image as input for Satirical Image Detection}
    \label{fig:image_det_pos}
\end{figure}

\begin{figure}[H]
    \centering
    \includegraphics[width=\columnwidth]{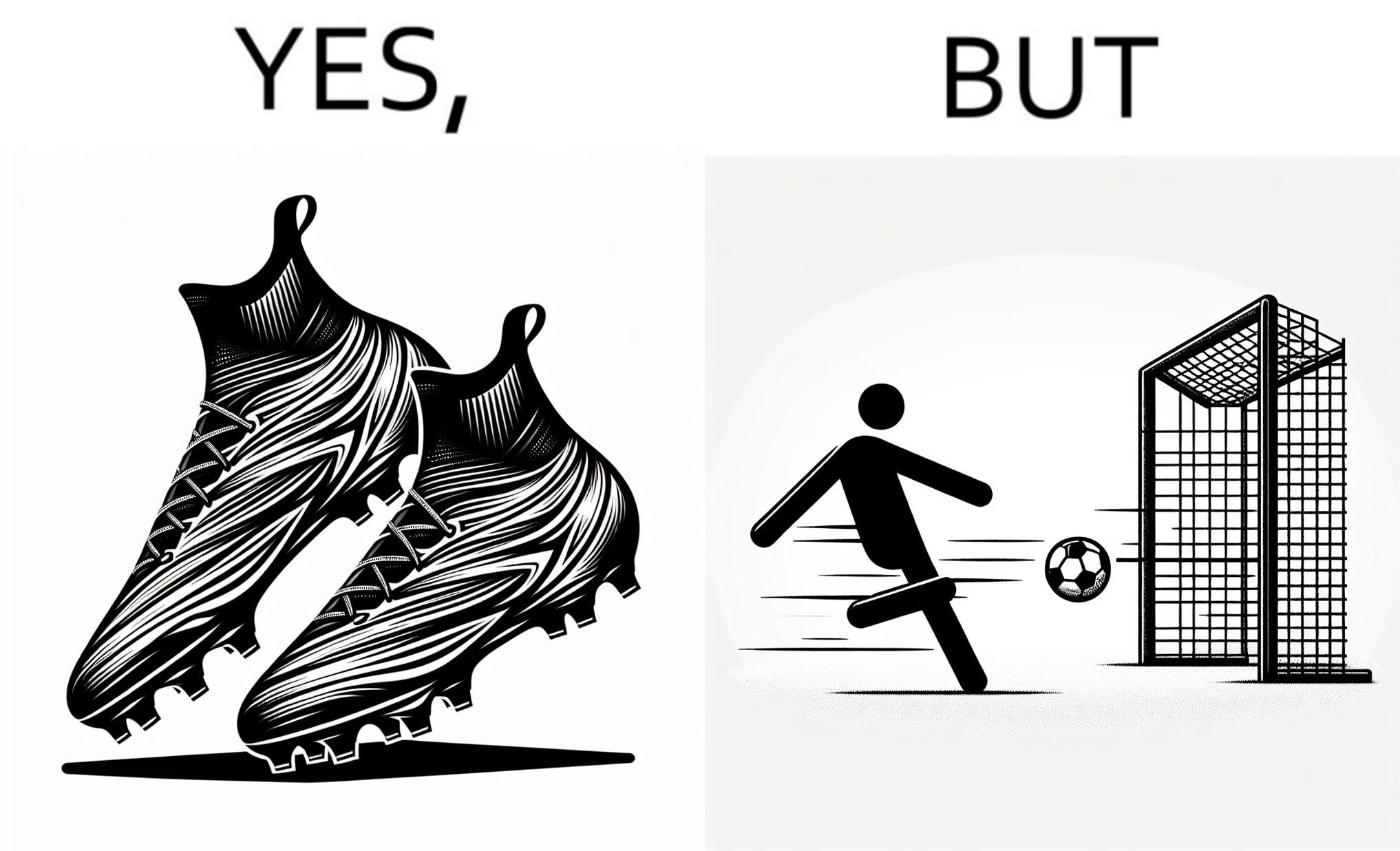}
    \caption{Example of a Non-Satirical Image as input for Satirical Image Detection}
    \label{fig:image_det_neg}
\end{figure}

Figures \ref{fig:image_det_pos} and \ref{fig:image_det_neg} are examples of satirical and non-satirical image inputs (for Satirical Image Detection) respectively. For each such image as input to the model, the aforementioned text prompt is used for Satirical Image Detection, and the output is either "Y" (predicting the image is satirical) or "N" (predicting the image is non-satirical).

\begin{figure}[H]
    \centering
    \includegraphics[width=\columnwidth]{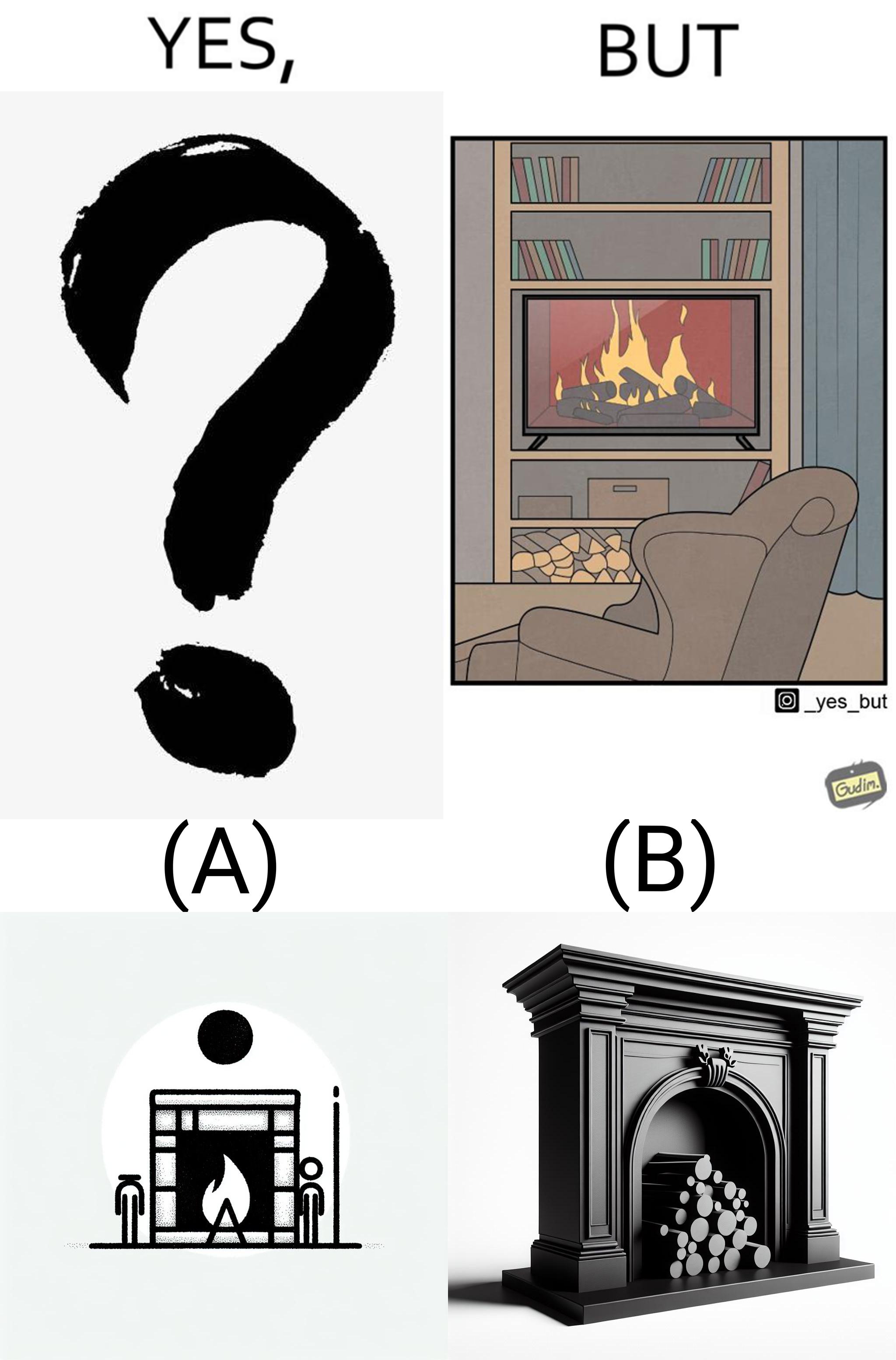}
    \caption{Example of an input image for Image Completion where the left sub-image is to be predicted [ground truth answer - (B)]}
    \label{fig:left_masked_completion}
\end{figure}

\begin{figure}[H]
    \centering
    \includegraphics[width=\columnwidth]{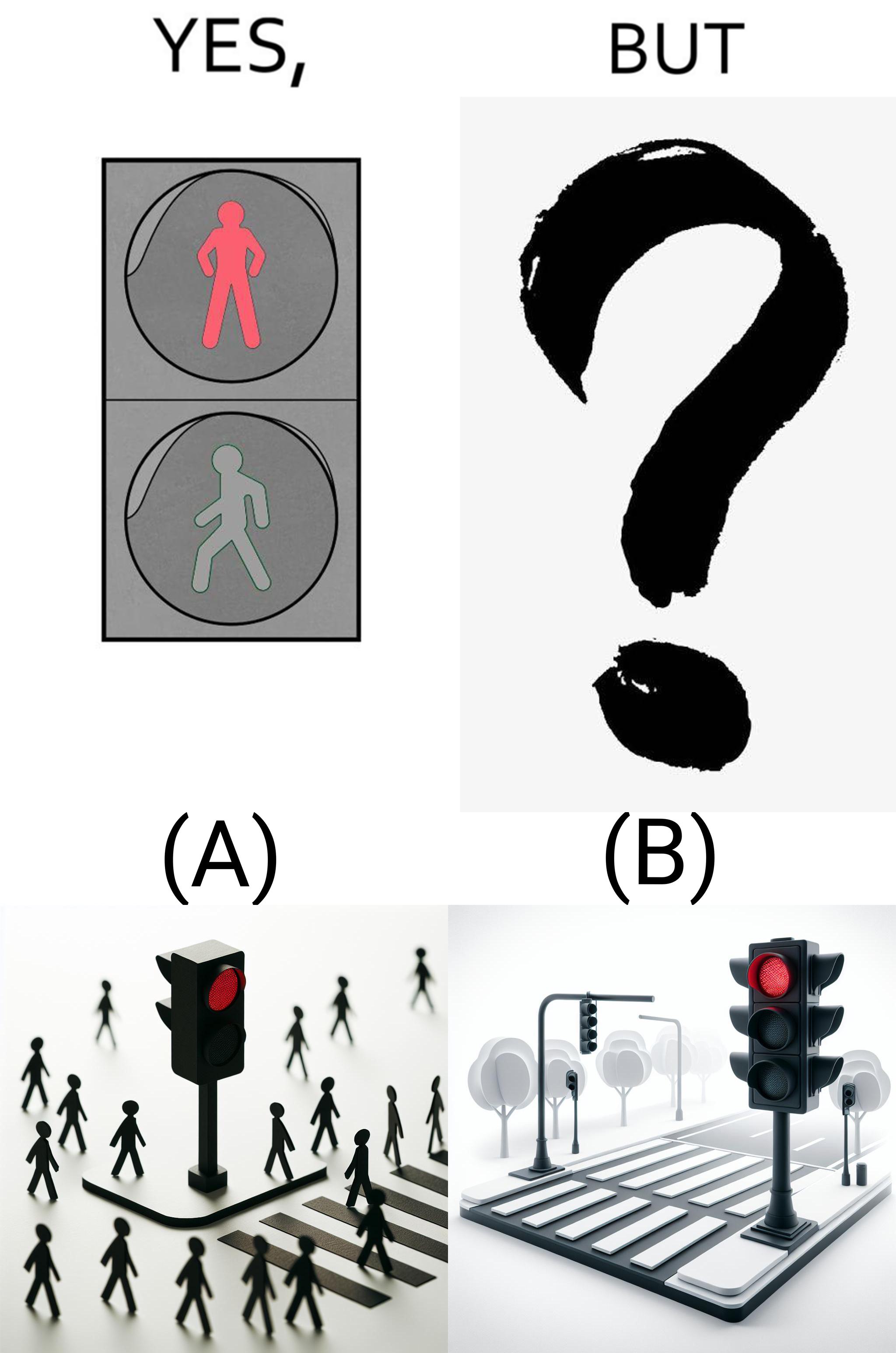}
    \caption{Example of an input image for Image Completion where the right sub-image is to be predicted [ground truth answer - (B)]}
    \label{fig:right_masked_completion}
\end{figure}

\noindent \textbf{Text Prompt for \textit{Satirical Image Completion}: }

\texttt{You are an AI expert in creating humour or satire. User gives you an image, and you have to make a choice "A" or "B".}

\texttt{Instructions: The image is a 2x2 table with the labels "yes" (top left), "but" (top right), "A" (bottom left), and "B" (bottom right). Either the "yes" cell or the "but" cell will have a question mark in it. Your job is to replace the question mark with either cell "A" or cell "B" so that the resulting [yes,but] pair is funny or satirical. Make a choice "A" or "B":}

\texttt{Output format: one character, exactly either "A" or "B".}

\noindent \textbf{Example Image Inputs for \textit{Satirical Image Completion}: }

Figures \ref{fig:left_masked_completion} and \ref{fig:right_masked_completion} are examples of input images for Image Completion where the left and right sub-image is to be predicted respectively. For each such image as input to the model, the aforementioned text prompt is used for Satirical Image Completion, and the output is either "A" or "B", denoting the sub-image predicted to come in place of the question mark in the input image.

\subsection{Evaluation Setup}
\subsection{Results}
\label{appendix:results}
% \usepackage{multirow}
% \usepackage{graphicx}
\begin{table}[H]
\centering
\resizebox{0.5\textwidth}{!}{%
\begin{tabular}{cc|cccccc}
\hline
\textbf{\begin{tabular}[c]{@{}c@{}}Stage of\\ Annotation\end{tabular}} &                                                                          & \textbf{Model}               & \textbf{BLEU}        & \textbf{ROUGE-L}     & \textbf{METEOR}      & \textbf{BERTScore}   & \textbf{\begin{tabular}[c]{@{}c@{}}Average\\ Score\end{tabular}} \\ \hline
\multirow{15}{*}{Stage 2}                                              & \multirow{5}{*}{\begin{tabular}[c]{@{}c@{}}LEFT\\ IMAGE\\ DESCRIPTION\end{tabular}}     & LLaVA                        & 0.0048               & 0.1471               & 0.2352               & 0.8588               & 0.3115                                                           \\
                                                                       &
                                                                       &
                                                                       \multicolumn{1}{l}{Kosmos-2}
                                                                       & 0.0113 &
                                                                       0.2444
                                                                       & 0.271
                                                                       & 0.881
                                                                       & 0.3519
                                                                       \\
                                                                       &                                                                                         & \multicolumn{1}{l}{MiniGPT4} & 0.0071               & 0.1519               & 0.2075               & 0.8352               & 0.3004                                                           \\
                                                                       &                                                                                         & GPT4                         & 0.0071               & 0.1207               & 0.2401               & 0.8506               & 0.3046                                                           \\
                                                                       &                                                                                         & Gemini                       & 0.0205               & 0.2513               & 0.2959               & 0.862                & 0.3574                                                           \\ \cline{2-8} 
                                                                       & \multirow{5}{*}{\begin{tabular}[c]{@{}c@{}}RIGHT\\ IMAGE\\ DESCRIPTION\end{tabular}}    & LLaVA                        & 0.0054               & 0.1465               & 0.225                & 0.8564               & 0.3083                                                           \\
                                                                       &
                                                                       &
                                                                       \multicolumn{1}{l}{Kosmos-2}
                                                                       &
                                                                       0.01
                                                                       & 0.2358
                                                                       & 0.2363
                                                                       & 0.88
                                                                       & 0.3405
                                                                       \\
                                                                       &                                                                                         & \multicolumn{1}{l}{MiniGPT4} & 0.0047               & 0.1427               & 0.183                & 0.8317               & 0.2905                                                           \\
                                                                       &                                                                                         & GPT4                         & 0.0052               & 0.114                & 0.2155               & 0.8435               & 0.2945                                                           \\
                                                                       &                                                                                         & Gemini                       & 0.0087               & 0.2259               & 0.2658               & 0.8655               & 0.3415                                                           \\ \cline{2-8} 
                                                                       & \multirow{5}{*}{\begin{tabular}[c]{@{}c@{}}WHYFUNNY\\ PROMPT\end{tabular}}              & LLaVA                        & 0.0098               & 0.1778               & 0.2329               & 0.8601               & 0.3201                                                           \\
                                                                       &                                                                                         & \multicolumn{1}{l}{Kosmos-2} &           0.0118           &        0.2103              &         0.1802             &              0.8708        &          0.3183                                                        \\
                                                                       &                                                                                         & \multicolumn{1}{l}{MiniGPT4} & 0.0013               & 0.1237               & 0.1671               & 0.8108               & 0.2757                                                           \\
                                                                       &                                                                                         & GPT4                         & 0.0033               & 0.1684               & 0.208                & 0.8607               & 0.3101                                                           \\
                                                                       &                                                                                         & Gemini                       & 0.0032               & 0.1973               & 0.2478               & 0.8425               & 0.3227                                                           \\ \hline
\multirow{15}{*}{Stage 3}                                              & \multirow{5}{*}{\begin{tabular}[c]{@{}c@{}}LEFT\\ IMAGE\\ DESCRIPTION\end{tabular}}     & LLaVA                        & 0.0062               & 0.1617               & 0.2327               & 0.8617               & 0.3156                                                           \\
                                                                       &
                                                                       &
                                                                       Kosmos-2
                                                                       & 0
                                                                       .0119
                                                                       &
                                                                       0.2284
                                                                       &
                                                                       0.2643
                                                                       &
                                                                       0.8769
                                                                       &
                                                                       0.3454
                                                                       \\
                                                                       &                                                                                         & MiniGPT4                     & \multicolumn{1}{l}{0.0061} & 0.1182 & 0.1806 & 0.8378 &   0.2857                                           \\
                                                                       &                                                                                         & GPT4                         & 0.0081               & 0.1427               & 0.2457               & 0.8554               & 0.313                                                            \\
                                                                       &                                                                                         & Gemini                       & 0.0145               & 0.2714               & 0.2714               & 0.8885               & 0.3615                                                           \\ \cline{2-8} 
                                                                       & \multirow{5}{*}{\begin{tabular}[c]{@{}c@{}}RIGHT\\ IMAGE\\ DESCRIPTION\end{tabular}}    & LLaVA                        & 0.0032               & 0.1567               & 0.23                 & 0.8603               & 0.3125                                                           \\
                                                                       &                                                                                         & Kosmos-2                     
                                                                       & 0.0083 
                                                                       & 0.2092 
                                                                       & 0.2499 
                                                                       & 0.8738
                                                                       & 0.3353                                             \\
                                                                       &                                                                                         & MiniGPT4                     & 0.0031 & 0.1092 & 0.1858 & 0.8231 &     0.2803                                         \\
                                                                       &                                                                                         & GPT4                         & 0.0044               & 0.133                & 0.2312               & 0.8509               & 0.3049                                                           \\
                                                                       &                                                                                         & Gemini                       & 0.009                & 0.2608               & 0.2593               & 0.8894               & 0.3546                                                           \\ \cline{2-8} 
                                                                       & \multirow{5}{*}{\begin{tabular}[c]{@{}c@{}}WHYFUNNY\\ PROMPT\end{tabular}}              & LLaVA                        & 0.0097               & 0.1795               & 0.2189               & 0.8588               & 0.3167                                                           \\
                                                                       &                                                                                         & Kosmos-2                     &         0.0093             &         0.1955             &          0.1943            &            0.8657          &                   0.3162                                               \\
                                                                       &                                                                                         & MiniGPT4                     & 0.0028               & 0.1499               & 0.1534               & 0.835                & 0.2853                                                           \\
                                                                       &
                                                                       & GPT4
                                                                       & 0.0027
                                                                       & 0.1466
                                                                       & 0.225
                                                                       & 0.8498
                                                                       & 0.306
                                                                       \\
                                                                       &                                                                                         & Gemini                       & 0.0081               & 0.1874               & 0.2321               & 0.8565               & 0.321                                                            \\ \hline
\multirow{15}{*}{Stage 4}                                              & \multirow{5}{*}{\begin{tabular}[c]{@{}c@{}}LEFT\\ IMAGE\\ DESCRIPTION\end{tabular}}     & LLaVA                        & 0.0064               & 0.1598               & 0.2388               & 0.8621               & 0.3168                                                           \\
                                                                       &
                                                                       &
                                                                       Kosmos-2
                                                                       & 0.0078
                                                                       & 0.2269 
                                                                       & 0.2662
                                                                       & 0.8756 
                                                                       & 0.3441
                                                                       \\
                                                                       &                                                                                         & MiniGPT4                     & 0.0063 & 0.1098 & 0.1836 & 0.8242 &         0.281                                     \\
                                                                       &                                                                                         & GPT4                         & 0.0073               & 0.1357               & 0.2378               & 0.8544               & 0.3088                                                           \\
                                                                       &                                                                                         & Gemini                       & 0.0233               & 0.2743               & 0.29                 & 0.8869               & 0.3686                                                           \\ \cline{2-8} 
                                                                       & \multirow{5}{*}{\begin{tabular}[c]{@{}c@{}}RIGHT\\ IMAGE\\  DESCRIPTION\end{tabular}} & LLaVA                        & 0.0057               & 0.1709               & 0.2437               & 0.8646               & 0.3212                                                           \\
                                                                       &                                                                                         & Kosmos-2                     
                                                                       & 0.0104
                                                                       & 0.2093 
                                                                       & 0.2452
                                                                       & 0.8738
                                                                       & 0.3347                                             \\
                                                                       &                                                                                         & MiniGPT4                     & 0.0043 & 0.1152 & 0.1866 & 0.8342 & 0.2851                                            \\
                                                                       &                                                                                         & GPT4                         & 0.007                & 0.136                & 0.2309               & 0.8529               & 0.3067                                                           \\
                                                                       &                                                                                         & Gemini                       & 0.0311               & 0.2856               & 0.2928               & 0.8923               & 0.3755                                                           \\ \cline{2-8} 
                                                                       & \multirow{5}{*}{\begin{tabular}[c]{@{}c@{}}WHYFUNNY\\ PROMPT\end{tabular}}              & LLaVA                        & 0.012               & 0.1809               & 0.2233               & 0.8596               & 0.319                                                           \\
                                                                       &                                                                                         & Kosmos-2                     & 0.0114 & 0.2007 & 0.2092 & 0.865 & 0.3216                                             \\
                                                                       &                                                                                         & MiniGPT4                     & 0.0014               & 0.1494               & 0.1519               & 0.8327               & 0.2838                                                           \\
                                                                       &
                                                                       & GPT4
                                                                       & 0.0024
                                                                       & 0.1444
                                                                       & 0.2211
                                                                       & 0.8488
                                                                       & 0.3042
                                                                       \\
                                                                       &                                                                                         & Gemini                       & 0.0101               & 0.1875               & 0.2355               & 0.8579               & 0.3227                                                           \\ \hline
\end{tabular}%
}
\caption{Evaluation of Satire Understanding on images curated at different Stages of annotation of \yesbut}
\label{tab:understanding}
\end{table}

% \usepackage{multirow}
% \usepackage{graphicx}
\begin{table}[H]
\centering
\resizebox{\columnwidth}{!}{%
\begin{tabular}{cc|cccc}
\hline
\textbf{}              & \textbf{}                 & \textbf{BLEU} & \textbf{ROUGE} & \textbf{METEOR} & \textbf{BERTScore} \\ \hline
\multicolumn{1}{c|}{E} & \multirow{3}{*}{LLaVA}    & 0.0094        & 0.1779         & 0.2337          & 0.8607             \\
\multicolumn{1}{c|}{M} &                           & 0.0113        & 0.1741         & 0.2235          & 0.8558             \\
\multicolumn{1}{c|}{D} &                           & 0.0167        & 0.1875         & 0.2434          & 0.8577             \\ \hline
\multicolumn{1}{c|}{E} & \multirow{3}{*}{Kosmos-2} & 0.0121        & 0.2126         & 0.1841          & 0.8714             \\
\multicolumn{1}{c|}{M} &                           & 0.0089        & 0.1969         & 0.1540          & 0.8683             \\
\multicolumn{1}{c|}{D} &                           & 0.0128        & 0.1894         & 0.1588          & 0.8604             \\ \hline
\multicolumn{1}{c|}{E} & \multirow{3}{*}{MiniGPT4} & 0.0014        & 0.1316         & 0.1696          & 0.8157             \\
\multicolumn{1}{c|}{M} &                           & 0.0012        & 0.1198         & 0.1412          & 0.7670             \\
\multicolumn{1}{c|}{D} &                           & 0             & 0.1284         & 0.1887          & 0.8240             \\ \hline
\multicolumn{1}{c|}{E} & \multirow{3}{*}{GPT4}     & 0.0028        & 0.1684         & 0.2091          & 0.8615             \\
\multicolumn{1}{c|}{M} &                           & 0.0080        & 0.1773         & 0.2124          & 0.8576             \\
\multicolumn{1}{c|}{D} &                           & 0             & 0.1355         & 0.1566          & 0.8498             \\ \hline
\multicolumn{1}{c|}{E} & \multirow{3}{*}{Gemini}   & 0.0028        & 0.1986         & 0.2507          & 0.8423             \\
\multicolumn{1}{c|}{M} &                           & 0.0031        & 0.1927         & 0.2232          & 0.8659             \\
\multicolumn{1}{c|}{D} &                           & 0.0170        & 0.1775         & 0.2512          & 0.7604             \\ \hline
\end{tabular}%
}
\caption{Effect of annotation difficulty on Satirical Understanding Performance using the WHYFUNNY prompt across several SOTA VL Models (E - EASY, M - MEDIUM, D - DIFFICULT)}
\label{tab:difficulty}
\end{table}

Table \ref{tab:difficulty} shows the effect of annotation difficulty on Satirical Understanding Performance. We infer that except for MiniGPT4, no other model performs semantically well (BERTScore) for difficult images. Also, in 12 out of 20 cases (5 VLMs x 4 metrics), VL Models fail to perform well for difficult images. Hence, there is a positive correlation between VLMs and Humans regarding what is difficult, especially from a semantic point of view.

% \usepackage{multirow}
% \usepackage{graphicx}
\begin{table}[H]
\centering
\resizebox{\columnwidth}{!}{%
\begin{tabular}{cc|cccc}
\hline
\textbf{}              & \textbf{}                 & \textbf{BLEU} & \textbf{ROUGE} & \textbf{METEOR} & \textbf{BERTScore} \\ \hline
\multicolumn{1}{c|}{Y} & \multirow{2}{*}{LLaVA}    & 0.0103        & 0.1843         & 0.2436          & 0.8612             \\
\multicolumn{1}{c|}{N} &                           & 0.0093        & 0.1720         & 0.2234          & 0.8591             \\ \hline
\multicolumn{1}{c|}{Y} & \multirow{2}{*}{Kosmos-2} & 0.0129        & 0.2079         & 0.1801          & 0.8698             \\
\multicolumn{1}{c|}{N} &                           & 0.0109        & 0.2124         & 0.1803          & 0.8716             \\ \hline
\multicolumn{1}{c|}{Y} & \multirow{2}{*}{MiniGPT4} & 0.0014        & 0.1354         & 0.1677          & 0.8093             \\
\multicolumn{1}{c|}{N} &                           & 0.0012        & 0.1257         & 0.1665          & 0.8121             \\ \hline
\multicolumn{1}{c|}{Y} & \multirow{2}{*}{GPT4}     & 0.0052        & 0.1786         & 0.2182          & 0.8653             \\
\multicolumn{1}{c|}{N} &                           & 0.0016        & 0.1594         & 0.1989          & 0.8567             \\ \hline
\multicolumn{1}{c|}{Y} & \multirow{2}{*}{Gemini}   & 0.0049        & 0.2056         & 0.2533          & 0.8364             \\
\multicolumn{1}{c|}{N} &                           & 0.0017        & 0.1902         & 0.2429          & 0.8479             \\ \hline
\end{tabular}%
}
\caption{Effect of the presence of text in images on Satirical Understanding Performance using the WHYFUNNY prompt across several SOTA VL Models (Y - Text is present in the image, N - Text is absent in the image)}
\label{tab:text_presence}
\end{table}

Table \ref{tab:text_presence} shows the effect of the presence of text in images on Satirical Understanding Performance. We see that in 15 out of 20 (5 VLMs x 4 metrics) cases, VL Models perform better on images with text vs. no text, suggesting that the absence of text in images makes it difficult to understand satire. This is supported by \citet{hayashi2024artwork}.

\noindent \textbf{Polos Metric}

% \usepackage{multirow}
% \usepackage{graphicx}
\begin{table}[H]
\centering
\resizebox{0.7\columnwidth}{!}{%
\begin{tabular}{c|ll}
\hline
\textbf{\begin{tabular}[c]{@{}c@{}}Stage of \\ Annotation\end{tabular}} & \multicolumn{1}{c}{\textbf{Model}} & \multicolumn{1}{c}{\textbf{Polos}} \\ \hline
\multirow{5}{*}{Stage-2}                                                & GPT4                               & 0.4823                             \\
                                                                        & Gemini                             & 0.5278                             \\
                                                                        & LLaVA                              & 0.4623                             \\
                                                                        & MiniGPT4                           & 0.3359                             \\
                                                                        & Kosmos-2                           & 0.4462                             \\ \hline
\multirow{5}{*}{Stage-3}                                                & GPT4                               & 0.4122                             \\
                                                                        & Gemini                             & 0.4849                             \\
                                                                        & LLaVA                              & 0.3721                             \\
                                                                        & MiniGPT4                           & 0.3161                             \\
                                                                        & Kosmos-2                           & 0.4468                             \\ \hline
\multirow{5}{*}{Stage-4}                                                & GPT4                               & 0.4146                             \\
                                                                        & Gemini                             & 0.488                              \\
                                                                        & LLaVA                              & 0.3758                             \\
                                                                        & MiniGPT4                           & 0.311                              \\
                                                                        & Kosmos-2                           & 0.4476      \\   \hline                   
\end{tabular}%
}
\caption{Evaluation of Satirical Understanding Performance across several SOTA VL Models using the WHYFUNNY Prompt and the image-based metric Polos.}
\label{tab:polos}
\end{table}

Table \ref{tab:polos} evaluates the Satirical Understanding Performance across several SOTA VL Models using the image-based Polos Metric. We can infer that all SOTA VL Models fail to perform well on the Polos Metric.

\noindent \textbf{Overall image descriptions (human-written and predicted by 5 SOTA Models)}

\begin{figure}[H]
    \centering
    \includegraphics[width=\columnwidth]{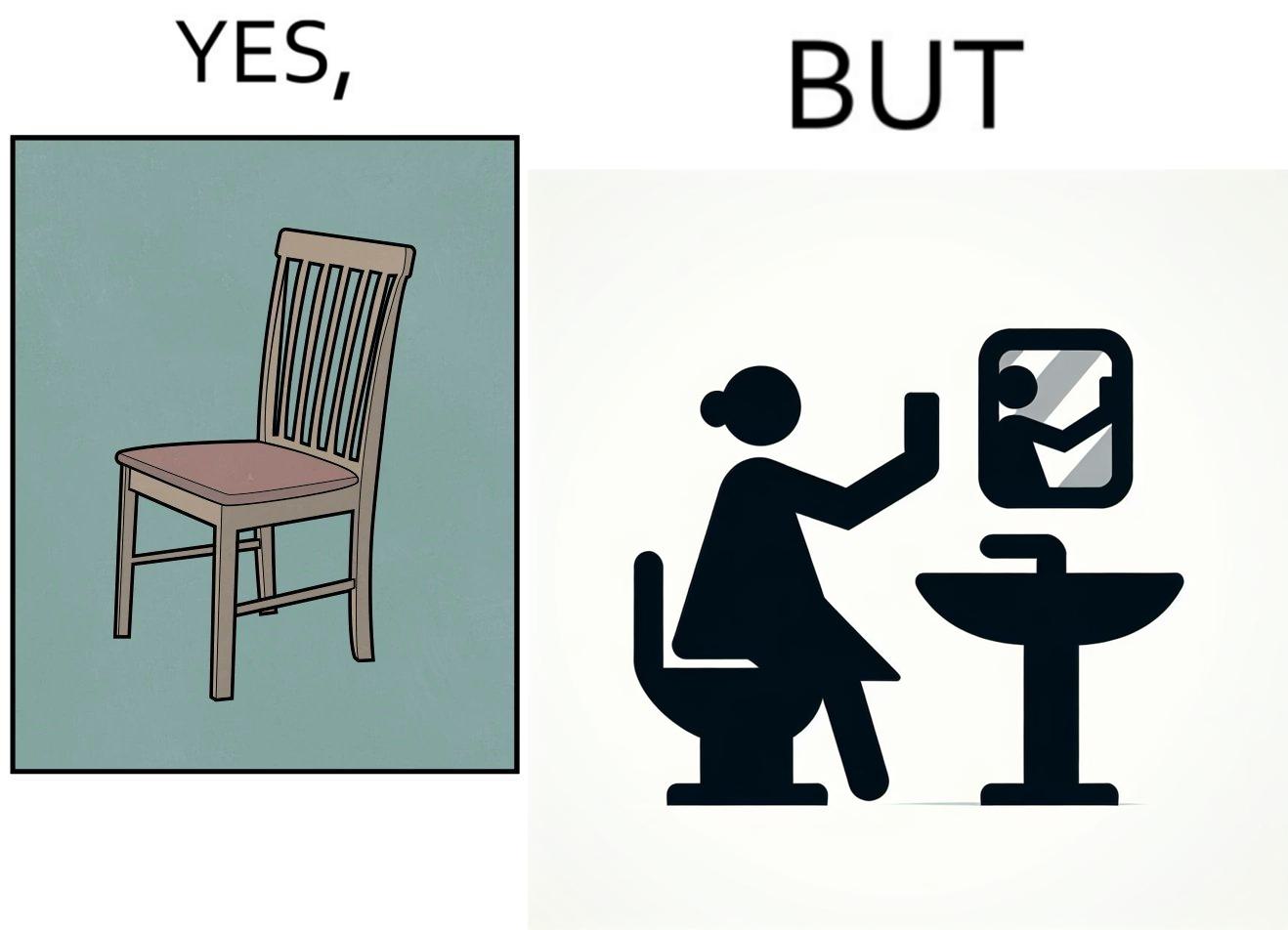}
    \caption{Example of a satirical image from \yesbut}
    \label{fig:appendix_yesbut}
\end{figure}

\begin{table*}[t]
\centering
\begin{tabular}{l|p{14cm}}
\hline
\textbf{} & \textbf{Overall Image Description} \\ \hline
\textbf{Human-written} & The image is ironical, as a woman is sitting by the sink taking a selfie using a mirror, while not using a chair that is actually meant for sitting. \\ \hline
\textbf{LLaVA} & The image is funny because it depicts a woman in a dress standing over a toilet in a bathroom, which is an unusual and unexpected setting for a woman dressed in a dress. This is a play on the common expectation that women should be dressed in more formal attire, such as a dress, when they are in a more polite or sophisticated environment, like a restaurant or a formal event. In this case, the woman's attire contrasts with the setting, creating a humorous and unexpected scene. \\ \hline
\textbf{Kosmos-2} & The image is funny/sarcastic because it depicts a woman sitting on a toilet, holding a checkmark, and saying "Yes, but." The image is meant to poke fun at the idea that women are often expected to perform tasks that are considered menial or unnecessary. The woman's actions are considered humorous because they are not typical for a woman to be doing, and the image is intended to make a point about the gender roles and expectations in the workplace. \\ \hline
\textbf{GPT4} & "The image you've provided plays on the typical format of public signs and presents a humorous twist. On the left side, you see a plain chair with the word ""YES"" above it, and on the right side, there's an icon-style figure of a person placing a voting ballot into a box with the word ""BUT"" above it.

The satirical element comes from the juxtaposition of the unequivocal ""YES"" with the more ambiguous ""BUT,"" suggesting a contrast between the straightforwardness of sitting in a chair versus the complexity or potential reservations one might have when casting a vote. It could be interpreted as a commentary on" \\ \hline
\textbf{MiniGPT4} & [TXT] "A cartoon depicting a person sitting at a desk with a toilet in the background. The person is holding a magnifying glass up to their face and looking at their reflection in the toilet bowl. The caption reads "yes, but. " \\ \hline
\textbf{Gemini} & The image is funny and satirical because it's a commentary on the way social media makes us feel. The woman in the image is sitting on a toilet but taking a selfie with a weird angle to try and make her look good for social media. In reality, we all know that no one looks good when taking a selfie on the toilet. The image is a reminder that social media is not always a true reflection of reality and that we should be careful not to compare ourselves to others. \\ \hline
\end{tabular}
\caption{Overall Image Descriptions (human-written and predicted by 5 SOTA Models) corresponding to Figure \ref{fig:appendix_yesbut}}
\label{tab:appendix_yesbut_desc}
\end{table*}

\begin{figure}[H]
    \centering
    \includegraphics[width=\columnwidth]{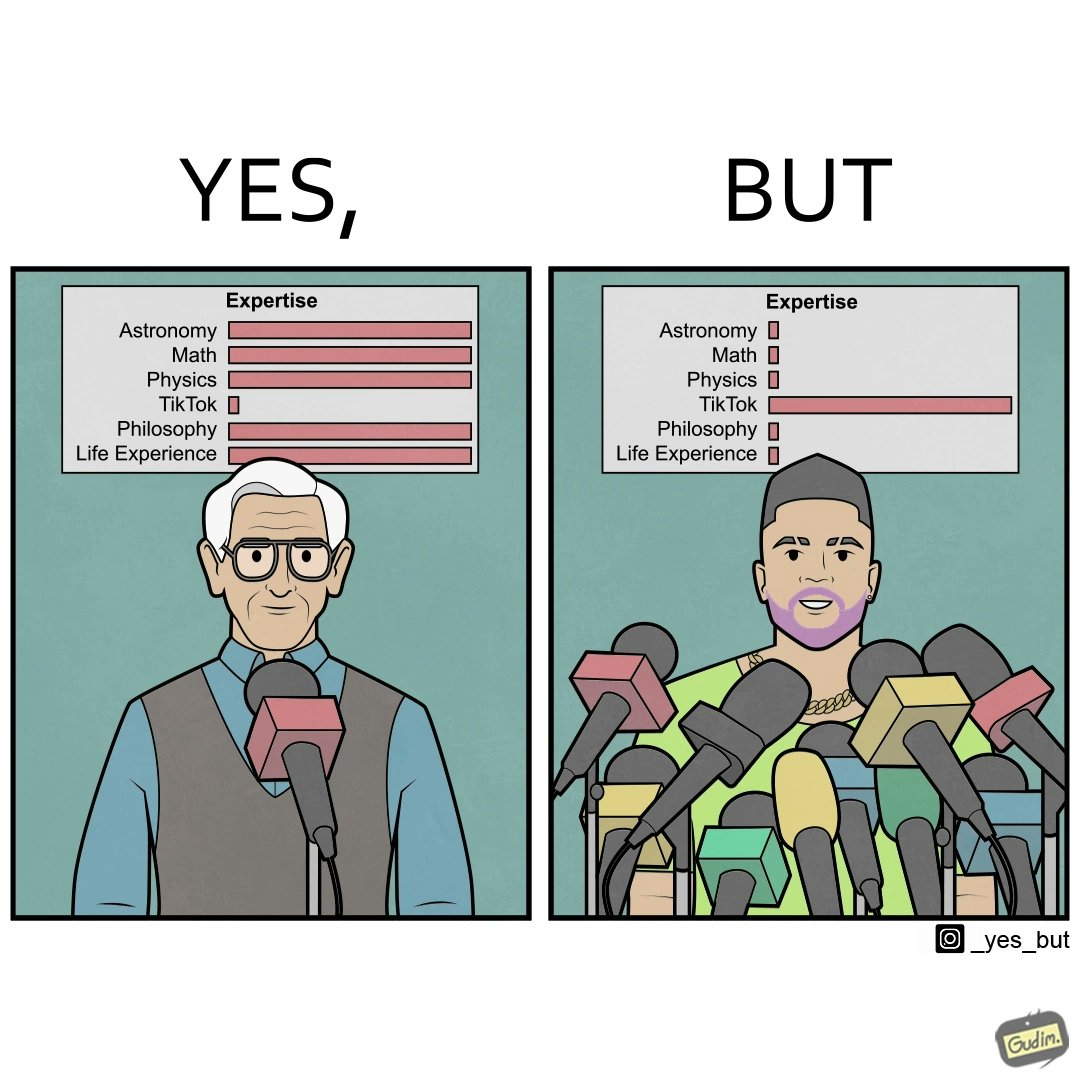}
    \caption{Example of a satirical image from \yesbut}
    \label{fig:appendix_yesbut_2}
\end{figure}

\begin{table*}[t]
\centering
\begin{tabular}{l|p{14cm}}
\hline
\textbf{} & \textbf{Overall Image Description} \\ \hline
\textbf{Human-written} & The image is satirical beacause it shows that people with more tiktok expertise are treated more importantly than the ones who are expert in all the other areas but tiktok. Here, the number of microphones a man speaks into is indicative of his importance. \\ \hline
\textbf{LLaVA} & The image is funny and satirical because it features a cartoon drawing of a man with a purple mohawk and glasses, surrounded by microphones, and the phrase "Expert, Astronomy" is displayed above him. This juxtaposition creates a humorous and ironic situation, as the man's appearance and occupation seem unrelated or even contradictory. The image likely intends to make a commentary on the absurdity of certain labels or assumptions people might make based on appearance or stereotypes. \\ \hline
\textbf{Kosmos-2} & The image is funny because it depicts an older man sitting in front of a row of microphones, asking a question to a young man who is wearing a mask. The older man is wearing glasses and has a beard, while the young man is not wearing a beard. The scene is set in front a podium, and the older man and the young boy are both wearing masks. This humorous scene is meant to poke fun at the idea that people often assume that older people are wiser \\ \hline
\textbf{GPT4} & The image is humorous because it shows an exaggerated contrast between two individuals' expertise levels across various disciplines, with one having significantly more experience and knowledge in each area. The satirical point lies in suggesting that sometimes people with lesser expertise might present themselves as equally or more knowledgeable, often leading to comical or ironic situations. \\ \hline
\textbf{MiniGPT4} & ? [INST \\ \hline
\textbf{Gemini} & The image is funny because it shows how people with a lot of expertise in a particular subject will talk about it in a very technical and complicated way, using a lot of jargon, while people with less expertise in the same subject will talk about it in a more general and relatable way. \\ \hline
\end{tabular}
\caption{Overall Image Descriptions (human-written and predicted by 5 SOTA Models) corresponding to Figure \ref{fig:appendix_yesbut_2}}
\label{tab:appendix_yesbut_desc_2}
\end{table*}

\begin{figure}[H]
    \centering
    \includegraphics[width=\columnwidth]{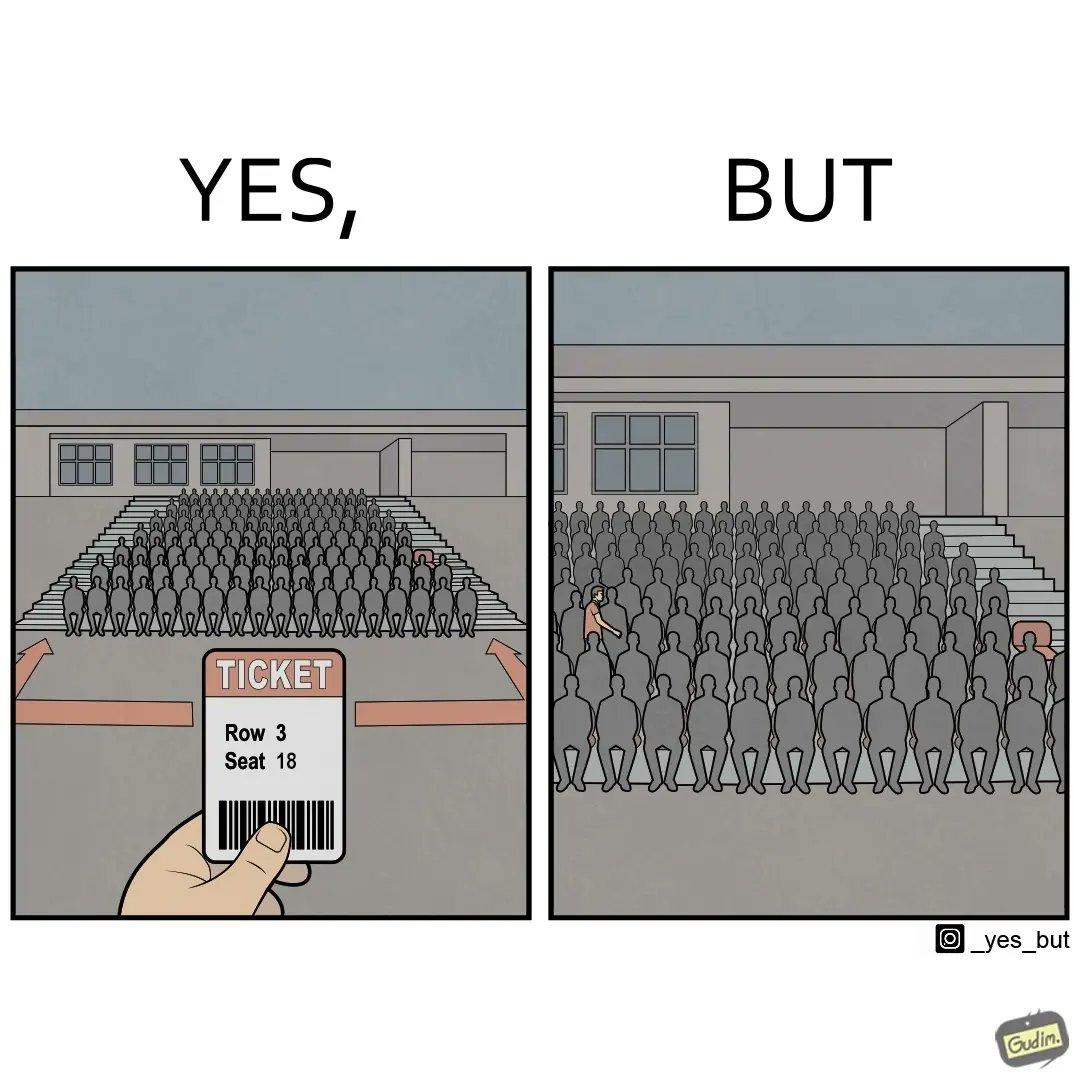}
    \caption{Example of a satirical image from \yesbut}
    \label{fig:appendix_yesbut_3}
\end{figure}

\begin{table*}[t]
\centering
\begin{tabular}{l|p{14cm}}
\hline
\textbf{} & \textbf{Overall Image Description} \\ \hline
\textbf{Human-written} & The image is funny, as the person has a ticket of a seat that is on the leftmost end of the gallery. However, the person has entered from the other end, and is going through a sea of seated people to get to the vacant seat. This would cause inconvenience to the people in the row, and would probably take a longer time to reach the seat. \\ \hline
\textbf{LLaVA} & The image is funny and satirical because it shows a person holding a ticket to a stadium with a large number of empty seats. The caption "Yes, but" is displayed above the image, suggesting that despite the empty seats, the event is still happening. This contrasts with the actuality of the event being poorly attended, which is further emphasized by the person being the only one in the image. The humor comes from the irony of the situation and the contrast between the expectation of a crowded event and the actuality of an empty stadium. \\ \hline
\textbf{Kosmos-2} & The image is funny because it depicts a theater full of people, but it's also satirical because it'll make you think about the audience's behavior during a movie. In the image, a person is holding a sign that says "Movies: Sold Out" and is pointing at the empty seats. This suggests that the theater is full, but the audience is not showing up. The person is trying to make a point about the lack of interest in the movie, and the empty chairs are a reflection of that. \\ \hline
\textbf{GPT4} & This image plays on the common frustration of attending an event with assigned seating, only to find that your seat is directly behind a pole or obstruction, blocking the view. It humorously exaggerates the issue by showing an entire audience perfectly aligned except for one obstructed seat, suggesting a comically unfortunate luck that the ticket holder has the only bad seat in the house. \\ \hline
\textbf{MiniGPT4} & Why is this image funny/satirical Title: Why is this image funny/satirical Author: AnonyMousse Date: March 22, 2023 Tags: comedy, satire Comments: The image shows a graph with two lines, one labeled "yes" and the other labeled "but". The graph appears to show a comparison between two different statements or ideas, but it is unclear what they are. The title and caption suggest that the image is meant to be humorous or satirical, but it is not clear \\ \hline
\textbf{Gemini} & The image is funny because the person holding the ticket looks very pleased with their seat assignment, while the image shows that there are plenty of open seats and the entire row is empty except for that one person. \\ \hline
\end{tabular}
\caption{Overall Image Descriptions (human-written and predicted by 5 SOTA Models) corresponding to Figure \ref{fig:appendix_yesbut_3}}
\label{tab:appendix_yesbut_desc_3}
\end{table*}

Tables \ref{tab:appendix_yesbut_desc}, \ref{tab:appendix_yesbut_desc_2}, and \ref{tab:appendix_yesbut_desc_3} contain the overall image descriptions (human-written and predicted by 5 SOTA Models) corresponding to Figures \ref{fig:appendix_yesbut}, \ref{fig:appendix_yesbut_2}, and \ref{fig:appendix_yesbut_3} respectively. We perform the following qualitative analysis on these 3 images as follows - 

\begin{itemize}
    \item Table \ref{tab:appendix_yesbut_desc} shows that no model gives correct reasoning behind why Figure \ref{fig:appendix_yesbut} is ironical, and each model makes different mistakes, For instance, GPT4 makes a blatant mistake of describing the right hand sub-image as a person placing a vote into a box. This shows the \textit{inability of SOTA VL Models to recognize objects properly when there is a mixture of artistic styles.}
    \item Figure \ref{fig:appendix_yesbut_2} shows that society views people as worthy today based on social media presence, rather than knowledge (worth is represented by number of microphones). Only GPT4 gives a close-to-correct reasoning. This shows the \textit{inability of SOTA VL Models to correlate objects in the image} (in this case, the number of microphones) \textit{to societal constructs} (in this case, worth).
    \item No VLM is able to decipher Figure \ref{fig:appendix_yesbut_3}, where the person looking for the assigned seat in a nearly-filled audience takes path of maximum resistance instead of going from the other side. This shows that \textit{SOTA VL Models are unable to understand miniature sketches of people/objects, as well as numbers} (the row, seat number here)
\end{itemize}

\begin{figure*}[t]
    \centering
\includegraphics[width=\textwidth]{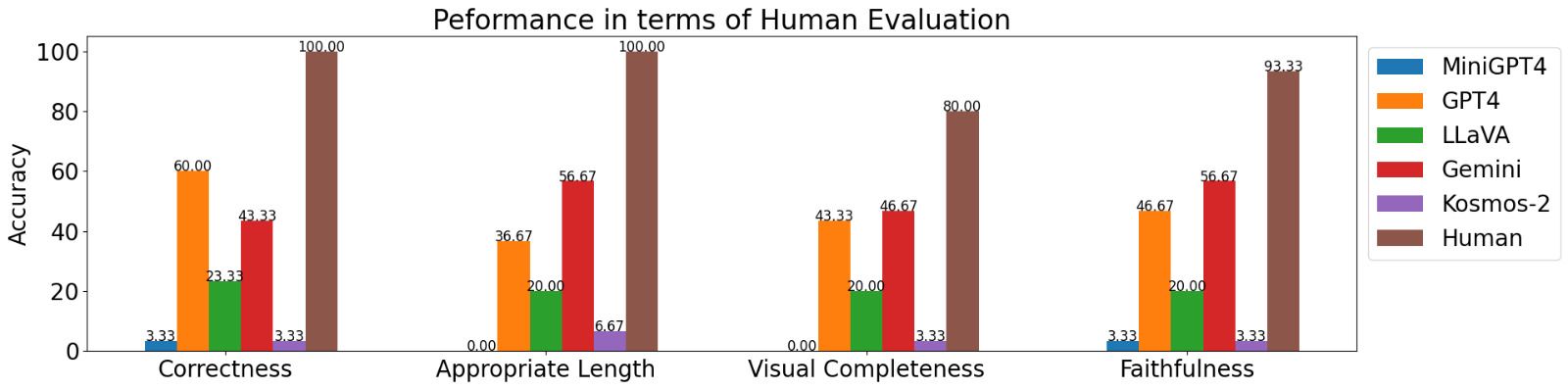}
    \caption{Results of Human Evaluation on the Satirical Image Understanding Task}
    \label{fig:human-eval}
\end{figure*}

Figure \ref{fig:human-eval} compares the overall image descriptions generated by 5 SOTA Models with ones written by human annotators based on human evaluation.

\section{Summary and Conclusion}

\end{document}